\documentclass[11pt]{article}

\usepackage[final]{acl}

\usepackage{times}
\usepackage{latexsym}

\usepackage[T1]{fontenc}

\usepackage[utf8]{inputenc}

\usepackage{microtype}
\newcommand{\change}[1]{{\footnotesize \textcolor{black!60}{#1}}}

\usepackage{inconsolata}

\usepackage{graphicx}

\usepackage{url}
\usepackage{multicol}
\usepackage{multirow}
\usepackage{textcomp}
\usepackage[dvipsnames]{xcolor}
\usepackage{subcaption}
\usepackage{makecell}
\usepackage{amsmath}
\usepackage{algorithm}
\usepackage{colortbl}
\usepackage{wrapfig}
\usepackage{cleveref}
\usepackage{amsfonts}       
\usepackage{nicefrac}       
\usepackage{microtype}      
\usepackage{xcolor}         
\usepackage{colortbl}
\usepackage{listings}
\usepackage{svg}

\usepackage{booktabs}
\usepackage{bm}

\lstdefinestyle{promptstyle}{
    basicstyle=\ttfamily\small,
    breaklines=true,
    breakatwhitespace=true,
    columns=fullflexible,
    frame=single,
    framerule=0.3pt,
    rulecolor=\color{black!30},
    backgroundcolor=\color{gray!5},
    showstringspaces=false,
    keepspaces=true,
    tabsize=2,
    numbers=none,
    xleftmargin=0pt,
    xrightmargin=0pt
}

\usepackage{amssymb}

\usepackage{tabularx}

\usepackage{enumitem}

\usepackage{algpseudocode}

\title{Ban\&Pick: Enhancing Performance and Efficiency of MoE-LLMs via Smarter Routing}

\author{
 \textbf{Yuanteng Chen\textsuperscript{1,2,3}},
 \textbf{Peisong Wang\textsuperscript{1,2}} \thanks{$\ $  Corresponding author.},
 \textbf{Yuantian Shao\textsuperscript{1,5}},
 \textbf{Nanxin Zeng\textsuperscript{2}},
\\
 \textbf{Chang Xu\textsuperscript{4}},
 \textbf{Jian Cheng\textsuperscript{1,2,3}} \footnotemark[1],
\\
 \textsuperscript{1}Institute of Automation, Chinese Academy of Sciences,
 \\
 \textsuperscript{2}School of Artificial Intelligence, University of Chinese Academy of Sciences,
 \\
 \textsuperscript{3}Zhongguancun Acedemy,
 \textsuperscript{4}School of Computer Science, University of Sydney,
 \\
 \textsuperscript{5}Nanjing University of Science and Technology
\\
 \small{
   \textbf{Correspondence:} \texttt{\{peisong.wang, jcheng\}@nlpr.ia.ac.cn}
 }
}

\begin{document}
\maketitle
\begin{abstract}
Mixture-of-Experts (MoE) has become a key architecture for scaling large language models. 
Recent fine‑grained MoE designs introduce hundreds of experts per layer, with multiple experts activated per token, enabling stronger specialization.  
However, during pre‑training, routers are optimized mainly for stability and robustness: they converge prematurely and enforce balanced usage, limiting the full potential of model performance and efficiency at inference.
In this work, we uncover two overlooked issues: (i) a few highly influential experts are underutilized due to premature and balanced routing decisions; and (ii) enforcing a fixed number of active experts per token introduces substantial redundancy.
Instead of retraining or redesigning MoE architectures, we introduce \textbf{Ban\&Pick}, a training-free, plug-and-play strategy for smarter routing. \textbf{Pick} discovers and reinforces key experts, a small group with outsized impact on performance, leading to notable accuracy gains. \textbf{Ban} further dynamically prunes redundant experts based on layer and token sensitivity, delivering faster inference.
Experiments on fine-grained MoE-LLMs (DeepSeek, Qwen3) across math, code, and general reasoning benchmarks demonstrate that Ban\&Pick delivers free performance gains and inference acceleration. For instance, on Qwen3-30B-A3B, it improves accuracy from 80.67 to 84.66 on AIME2024 and from 65.66 to 68.18 on GPQA-Diamond, while accelerating inference by 1.25× under the \texttt{vLLM}.
\end{abstract}

\section{Introduction}

Large language models (LLMs) have achieved remarkable progress \citep{DBLP:conf/iclr/ZhouWLSLQLJSZ024}, yet the ever-growing demand for higher capacity presents significant challenges for efficiency. As a promising solution, Mixture-of-Experts (MoE) architectures scale parameter count without proportionally increasing computation, by activating a small subset of experts for each input \citep{artetxe-etal-2022-efficient}.

Recent advances have pushed MoE towards a fine-grained design \citep{dai-etal-2024-deepseekmoe}, where each layer contains a large number of relatively small experts. For example, DeepSeek-v2.5 \citep{deepseekai2024deepseekv2strongeconomicalefficient} employs 160 experts per layer, while Qwen3-MoE \citep{yang2025qwen3technicalreport} adopts 128 experts per layer. Unlike coarse-grained MoE \citep{jiang2024mixtral}, which typically consists of a few large experts derived from the same base model, fine-grained MoE initializes each expert independently and randomly. 
This design promotes specialization: experts gradually develop distinct skills, excelling in areas such as math, code, or general reasoning. Such strong differentiation suggests that fine-grained MoE has the potential to leverage expert diversity more effectively.

\begin{figure*}[t]
    \centering
    \includegraphics[width=\textwidth]{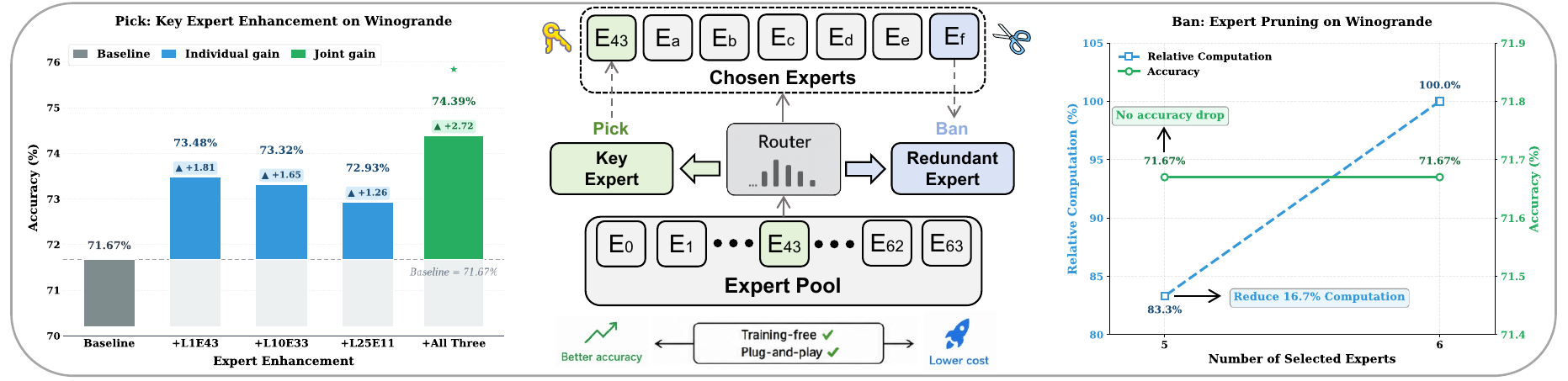}
    \caption{The intuition and empirical findings on expert utilization. 
    Left (\textbf{Pick}): forcibly activating key experts (e.g., E43) improves accuracy on Winogrande~\citep{sakaguchi2021winogrande}. 
    Right (\textbf{Ban}): reducing experts from 6 to 5 cuts 16.7\% computation without accuracy loss. And we provide additional results on more datasets in Appendix~\ref{app:more_motivation}.}
    \label{fig:motivation}
    \vspace{-10pt}
\end{figure*}

However, we argue that a critical but overlooked phenomenon exists: \textbf{the routing strategies learned during pre-training cannot fully exploit this specialization}. Specifically, we empirically discover that current routing strategies inadvertently limit the potential to leverage the most impactful experts while simultaneously introducing redundancy where many selected experts contribute minimally to the final output.
Instead of attempting to redesign training objectives or expert architectures (related work is discussed in detail in Appendix~\ref{app:related_work}), we aim to unlock this untapped potential efficiently and effectively during inference.

Delving into this question, we surprisingly find that within the most frequently selected experts of fine‑grained MoE models, a small fraction plays an outsized role: simply forcing these experts to be chosen for every input can noticeably boost accuracy on certain datasets. Motivated by this striking observation, we conduct an empirical analysis of expert usage across domains and discover that each domain consistently relies on its own set of frequently selected experts, which we refer to as domain‑specialized experts. 
Yet, a closer examination further shows only a tight subset of these experts has a decisive impact on the final logits distribution, while the influence of the others remains marginal. We term these most critical ones key experts. Our \textbf{Pick} module is designed to reinforce their influence during routing, 
getting performance gains by emphasizing key experts.

In parallel, we aim to alleviate redundancy in expert activation. Under the constraint that a fixed number of experts must be selected for each token, many activated experts contribute little to the final output, reflecting the redundancy introduced by less relevant experts. Our \textbf{Ban} module jointly considers layer sensitivity and token sensitivity to assess the redundancy in expert selection, and applies dynamic pruning of experts during inference. In this way, Ban reduces unnecessary computation and achieves significant inference acceleration with only minor accuracy loss.



Attempting to bridge the gap between routing strategies learned during pre-training, which prioritizes stability and balance, and optimal inference-time expert utilization, Pick and Ban empower truly impactful experts to contribute more while reducing participation from low-contribution experts:
\begin{itemize}[leftmargin=12pt, topsep=2pt, itemsep=1pt]
    \item \textbf{Pick} amplifies excellence by identifying experts that demonstrate superior performance on specific domains and strategically increasing their influence during inference. 
    
    \item \textbf{Ban} eliminates inefficiency by dynamically excluding experts that contribute minimally at inference, reducing computational costs while keeping essential capabilities.
\end{itemize}
Together, we introduce Ban\&Pick, a unified framework that optimizes expert utilization in inference to improve both the accuracy and efficiency of fine-grained MoE without retraining or architectural changes. Evaluated on five widely used datasets spanning math, code, and general reasoning, the results demonstrate that on the Qwen3 series models, applying Pick alone brings an average 2.62\% performance improvement. By further incorporating Ban, the joint Ban\&Pick strategy achieves an average 1.28\% improvement while simultaneously accelerating inference by 1.25× under \texttt{vLLM}.



\section{Empirical Findings on Routing Limitations in Fine-grained MoE}
\label{sec:motivation}

\subsection{Theoretical Insights from Pre-training}

Recent evidence from OLMoE \citep{muennighoff2025olmoeopenmixtureofexpertslanguage} and OpenMoE \citep{xue2024openmoe} suggests that during pre-training, routers converge much earlier than the experts. In particular, OLMoE reports that after only 1\% of pre-training, around 60\% of the routing decisions have already saturated, while OpenMoE finds that by 40\% of pre-training, about 80\% of the routing has stabilized. Such early fixation implies that routing decisions are largely determined before the experts have sufficiently matured. This premature convergence, combined with the widely used auxiliary balancing loss, fundamentally limits optimal expert utilization.

The balancing loss, introduced to prevent training collapse onto a few dominant experts, is typically formulated as $\mathcal{L}_{\text{expert}} = \alpha \sum_i f_i \cdot p_i$, where $f_i$ is the activation frequency and $p_i$ is the average routing probability of expert $i$ \citep{lepikhin2020gshardscalinggiantmodels}. When an expert receives both high routing probability and high actual usage, the product $f_i p_i$ becomes large and is penalized. While effective at ensuring balanced loads, this mechanism inherently conflicts with expert specialization \citep{wang2024auxiliary}—it discourages tokens from concentrating on the most competent experts, which is also argued by prior work~\citep{guo2025advancingexpertspecializationbetter}. Fine-grained MoE partially alleviates this tension by increasing the number of experts, allowing some specialization to emerge. However, as we will show in Section~\ref{sec:expert-specialization}, even strongly specialized experts rarely exceed $\sim$50\% activation frequency, indicating the balancing loss still constrains how dominant any single expert can become.

The interplay between premature routing and balancing constraints creates a two-stage mechanism that underlies key expert under-utilization. \textit{In the early stage}, before routing prematurely converges, the router learns a relatively stable domain-to-expert mapping, but the balancing loss caps how high any single expert's frequency can become. Once routing stabilizes, \textit{the remainder of training} (which constitutes the majority) continues to update expert parameters: frequently selected experts receive domain-specific data and become increasingly specialized. Crucially, however, their activation frequencies remain frozen at the earlier balance-constrained levels. \textbf{As a result, key experts that have evolved to be highly impactful are still activated at frequencies determined before their specialization matured—creating a fundamental mismatch between their true competence and actual utilization.}

This implies that the current pre-training paradigm makes a trade-off: in exchange for training stability, it locks routers into decisions before experts have fully matured. This presents an opportunity from an inference perspective:

\textbf{\textit{Can we recover this lost potential—amplifying the influence of the most impactful experts while removing redundancy, to boost performance and accelerate inference?}}


\subsection{Expert Utilization Inefficiencies and Optimization Potential}
To validate the inefficiencies in expert utilization and better understand the optimization potential, we conducted two simple yet revealing experiments with the Deepseek‑V2‑Lite‑Chat (activates 6 experts out of 64) on Winogrande (more results are provided in Appendix~\ref{app:more_motivation}). Figure~\ref{fig:motivation} demonstrates how we adjust expert utilization and its effects.

\textbf{(1) Underutilization of the most impactful experts:} we examined expert selection frequency on the C4 calibration set, identified the top-3 most frequently selected experts in each layer, and then conducted a grid search: each of these experts was manually added in addition to the original routing (E43 in Figure~\ref{fig:motivation}), and the resulting accuracy on Winogrande benchmark was recorded. The left part highlights a few experts contribute disproportionately: enhancing Layer‑1 Expert‑43 improved accuracy by +1.81\%, L10E33 by +1.65\%, and L25E11 by +1.26\%. When these three were jointly enhanced, accuracy increased by +2.72\% (71.67\% $\rightarrow$ 74.39\%), while most other high-frequency experts had negligible or noisy effects, indicating that certain impactful experts are insufficiently leveraged in the original routing strategy. We refer to these experts as \textbf{key experts}.

\textbf{(2) Computational waste from redundant experts:} we directly reduced the number of selected experts per token from 6 to 5 and measured the accuracy. Remarkably, as shown in the right side of Figure~\ref{fig:motivation}, this pruning reduced computation by 16.7\% and preserved full accuracy on Winogrande (71.67\%), demonstrating redundancy in the original expert utilization.
To rule out the possibility that this observation is dataset-specific, we provide consistent results across datasets in Appendix~\ref{app:more_motivation}.

Together, these experiments validate the inefficiencies in expert utilization predicted by pre-training analysis, and support the intuition of our methods—Pick and Ban, respectively, which aim to optimize expert selection beyond pre-training constraints for better inference-time performance. This empirical study motivates us to design a unified framework that can pick key experts while banning redundant ones, aiming at improving both accuracy and efficiency.

\section{Pick}
\label{sec:pick}
In this section, we begin with an analysis of expert specialization in fine‑grained MoE models to verify the existence of domain-specialized experts, which forms the foundation for key experts to effectively function. Building on this, we then introduce our Pick method in two steps: (i) identifying key experts from the set of domain‑specialized experts, and (ii) designing strategies to enhance key experts.

\subsection{Expert Specialization in Fine-Grained MoE}
\label{sec:expert-specialization}
\begin{figure*}[t]
    \centering
    \includegraphics[width=\textwidth]{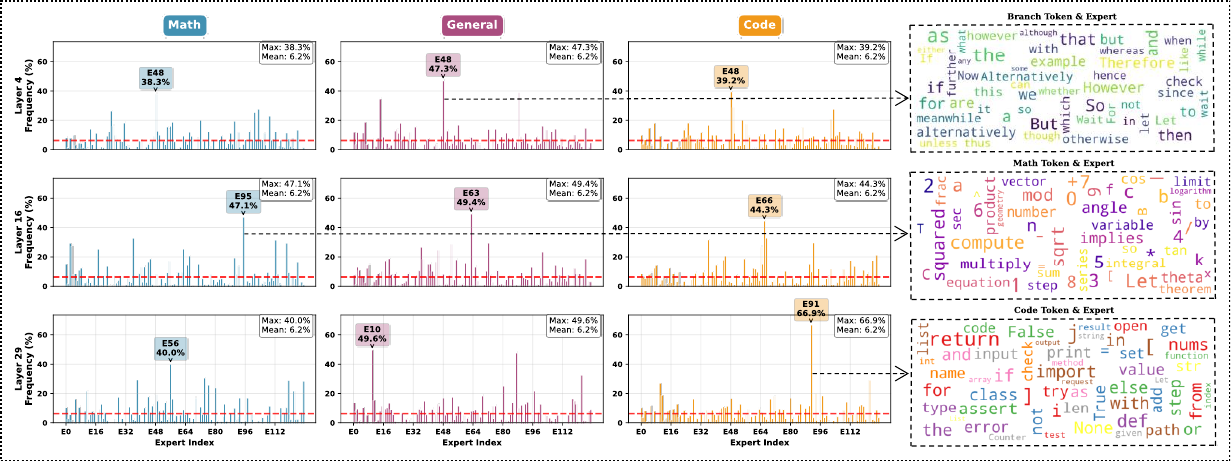 }
    \caption{Expert specialization in fine-grained MoE (Qwen3‑30B‑A3B). 
    Left: expert usage frequency across three tasks (math, general, code). 
    Right: token word clouds for 3 high-frequency experts.}
    \label{fig:expert_special}
    \vspace{-10pt}
\end{figure*}
First, we take a closer look at the degree of expert specialization in fine‑grained MoE models. Specifically, we analyze the Qwen3‑30B‑A3B model and record expert usage frequencies on task-related benchmarks: math tasks (GSM8K + Math-500), code tasks (MBPP~\citep{austin2021program} + Livecodebench), and a general task (GPQA-Diamond). The statistics cover all tokens in both the prefill and decode stages. For each type of task, we plot the per‑layer expert usage frequency and show 3 representative layers in Figure~\ref{fig:expert_special}. 

Our analysis reveals that in each layer there exist experts whose usage frequency is significantly higher than average--often up to 6 times higher. Moreover, in most layers these high‑frequency experts vary across tasks. 
For instance, L16E95 is frequently activated only on math tasks, whereas L29E91 is almost exclusively activated on code tasks with an extremely high activation rate of 66.9\%. In contrast, a small number of experts such as L4E48 remain highly active across 
all tasks.

These task‑dependent high‑frequency experts suggest the existence of domain‑specialized experts. To validate this hypothesis, we further examine the functional roles of these experts by analyzing which tokens are most often routed to them. Taking the three aforementioned experts as examples, we visualize the tokens they frequently process using word clouds.
The results show that the math‑specialized expert L16E95 is strongly associated with digits, operators, and math terminology; the code‑specialized expert L29E91 is associated with programming tokens; and the general expert L4E48 is closely tied to branch tokens \citep{zhong2025hybrimoehybridcpugpuscheduling}, which play a critical role in chain‑of‑thought across diverse tasks.

These observations provide empirical evidence that expert specialization in fine‑grained MoE models is highly pronounced. Different tasks rely heavily on different domain‑specialized experts, while a small subset of general experts remain consistently important across domains.

\subsection{Picking Key Experts from Domain‑Specialized Experts}
\label{sec:pick-key-experts}

As shown previously, almost every layer contains domain‑specialized experts that are activated with notably high frequency. The key issue, however, is whether all of these experts contribute equally to model’s performance, or only a small subset is truly essential. To investigate this, we evaluate the importance of each expert by measuring the shift in the logits distribution when it is removed.

Concretely, taking math tasks as an example, we use the domain‑specialized experts identified in previous analysis as candidate experts in each layer. We then run the model on 1000 randomly sampled instances from MathQA \citep{amini-etal-2019-mathqa} dataset. For each sample, we record the output logits of the original model, and then repeat the inference while pruning one candidate expert at a time. The perturbation of the output distribution is quantified using the KL divergence between the probability distributions of the Top‑1000 tokens from the original and pruned models.

\begin{figure}[tbh]
    \centering
    \includegraphics[width=\linewidth]{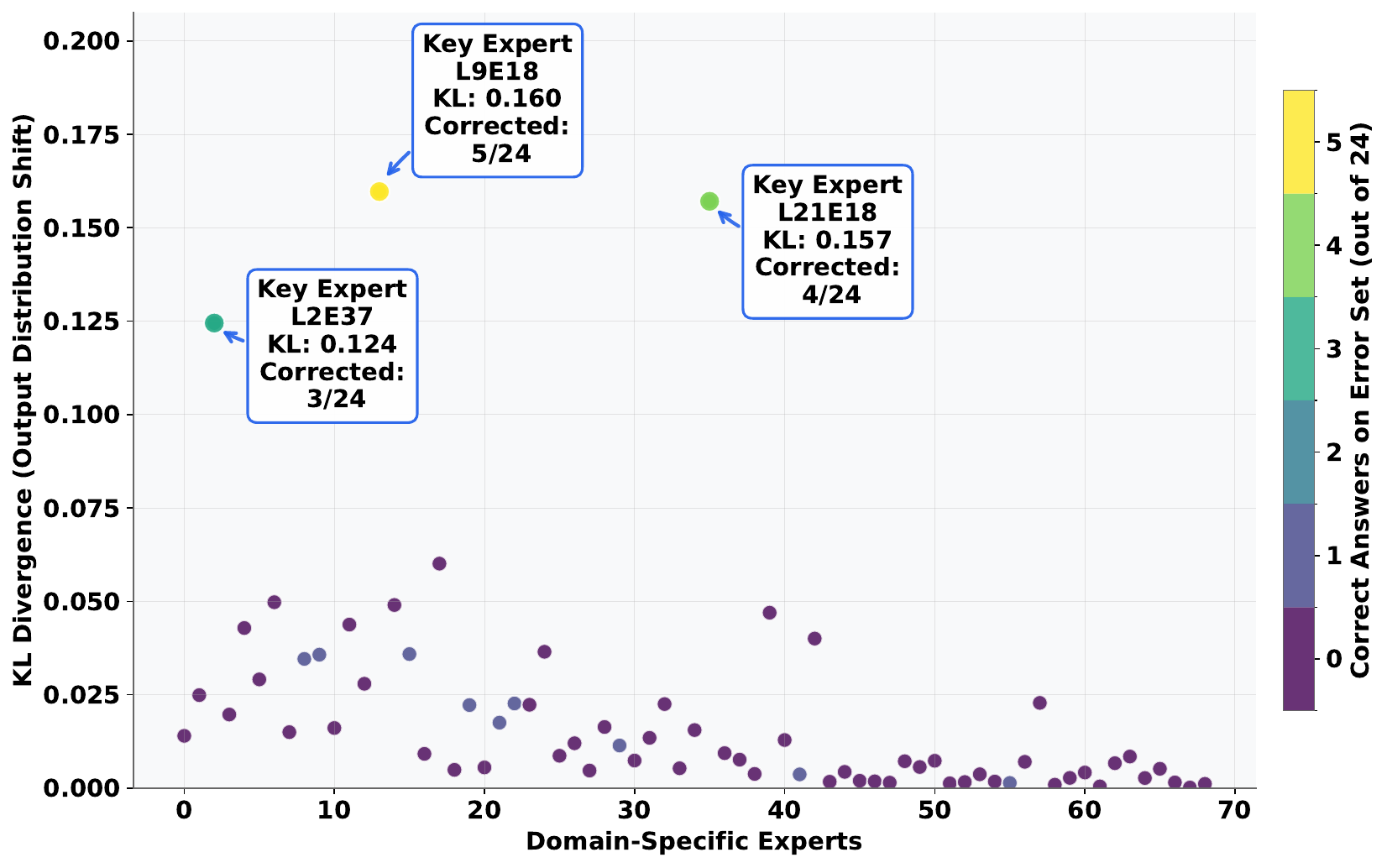}
    \caption{Impact of pruning (measured by KL divergence, left axis) and enhancing (by accuracy gain, right axis) for domain-specialized experts.}
    \label{fig:kl_expert}
    \vskip -0.1in
\end{figure}

The results are summarized in Figure~\ref{fig:kl_expert}. Surprisingly, despite their high activation frequency, most domain‑specialized experts cause only negligible shifts in the output distribution when pruned. In contrast, a small number of experts stand out, with three of them inducing particularly large changes. Among these, expert L9E18 is the most striking, leading to a divergence of about $0.16$. These findings suggest that only a few domain‑specialized experts truly have a decisive influence on the model’s behavior, and we hypothesize that they are the key experts we aim to identify.  


To test this hypothesis, we construct a failure set consisting of the Math‑500 \citep{lightman2023lets} problems that the base model originally answers incorrectly (24 in total). For each candidate expert, we forcibly include it during routing and then re‑evaluate accuracy on this failure set. As shown in Figure~\ref{fig:kl_expert} (right vertical axis), only those experts that induce substantial distribution shifts also yield meaningful accuracy gains when enhanced, whereas others contribute little to no improvement. In particular, expert L9E18, which caused the largest divergence, also increases the number of correct answers by 5 out of 24 when included. This provides strong evidence of the close correspondence between distribution shifts and functional importance, and offers a principled way to distinguish key experts from the many domain‑specialized experts.
We provide more implementation details on the selection criterion, calibration set choices, and KL divergence computation, together with complete results in Appendix~\ref{app:identify-key-experts}.

\subsection{Enhancing Key Experts}
\label{sec:enhancing}

\begin{figure}[tbh]
    \centering
    \includegraphics[width=\linewidth]{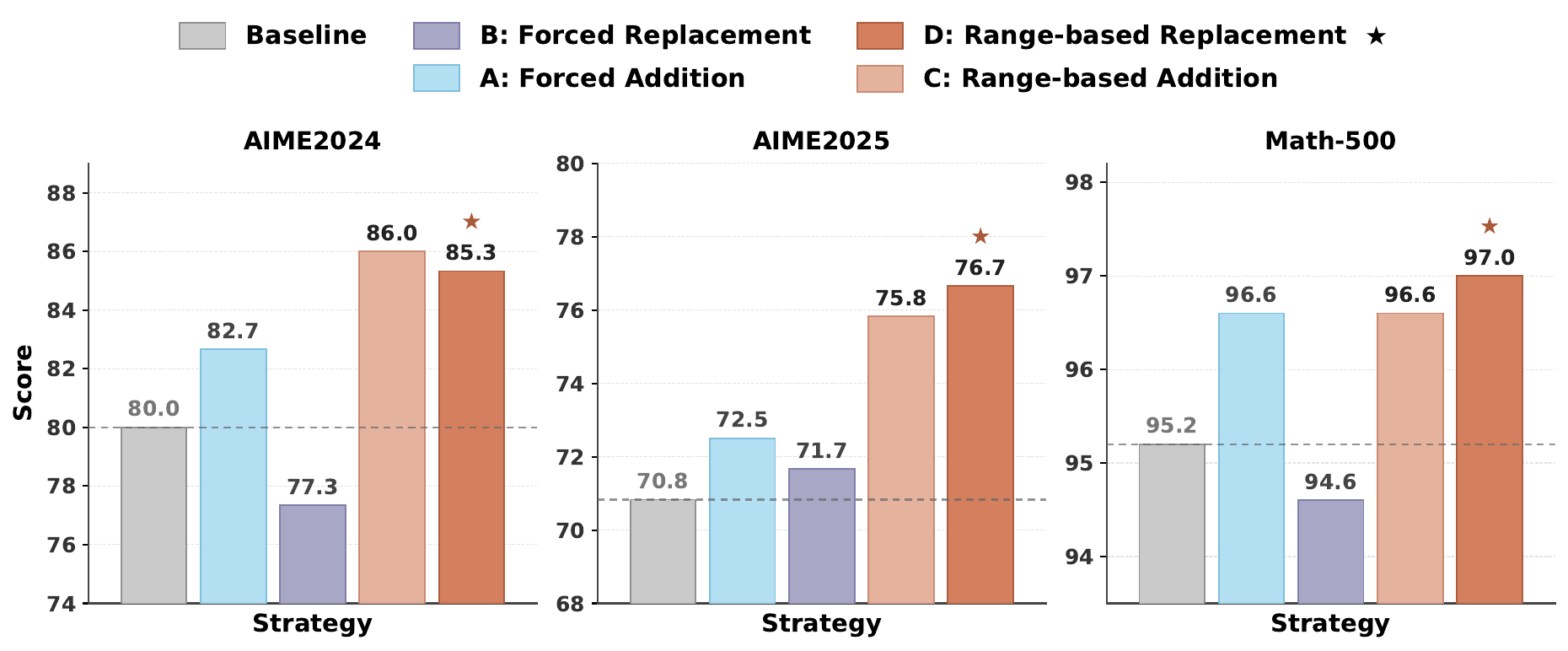}
    \caption{Comparison of four designed enhancement methods for key experts, evaluated by accuracy on three widely used math benchmarks.}
    \label{fig:expert_enhance}
    \vskip -0.1in
\end{figure}


After identifying the key experts, the remaining challenge is to determine how to leverage them most effectively. We take math tasks as a running example and design four enhancement strategies:

    
    
    
    

\begin{enumerate}[leftmargin=15pt, label=\textbf{\Alph*.}, itemsep=1pt, topsep=3pt]
    \item \textbf{Forced Addition:} Whenever a key expert is not selected, it is added alongside chosen experts.
    
    \item \textbf{Forced Replacement:} Similar to forced addition, but the key expert replaces the selected expert with the lowest routing weight, maintaining the total number of selected experts.
    
    \item \textbf{Range-based Addition:} If a key expert is not selected (not in the top-$k$ candidates) but appears within the router's top-$2k$ candidates, it is forcibly added.
    
    \item \textbf{Range-based Replacement:} Under the same condition as range-based addition, the key expert replaces the selected expert with the lowest routing weight.

\end{enumerate}


We evaluate these methods on AIME2024, AIME2025, and Math‑500. For the two AIME datasets, results are averaged over five runs to mitigate small‑sample variance. Figure~\ref{fig:expert_enhance} summarizes the outcomes. Overall, strategy A brings consistent but modest gains. Strategy B, which forces replacement regardless of probability, is unstable and even reduces accuracy on AIME2024 and Math‑500, indicating that overly aggressive intervention can be harmful. By contrast, strategies C and D, which enhance only when the key expert is already close to the router’s top candidates, deliver the strongest improvements across all benchmarks. The difference between C and D is minor, but since D avoids expanding the number of active experts, it is preferable in practice and used in experiments. 

\section{Ban}
\label{sec:dynamic-pruning}

Our dynamic expert pruning method is motivated by two key observations:  
1) Different layers in an MoE model exhibit highly diverse sensitivity to expert pruning and thus should be treated differently.  
2) During reasoning, the token-wise routing distributions also vary significantly: when most of the routing weight is concentrated on a few top experts, expert pruning causes minimal loss; in contrast, when the weights are more evenly distributed, expert pruning becomes more harmful.  

In the following, we first use Qwen3-30B-A3B as an example model to illustrate the differences in layer sensitivity and token sensitivity separately, and then introduce our dynamic pruning method. 


\begin{figure}[tbh]
    \centering
    \includegraphics[width=\linewidth]{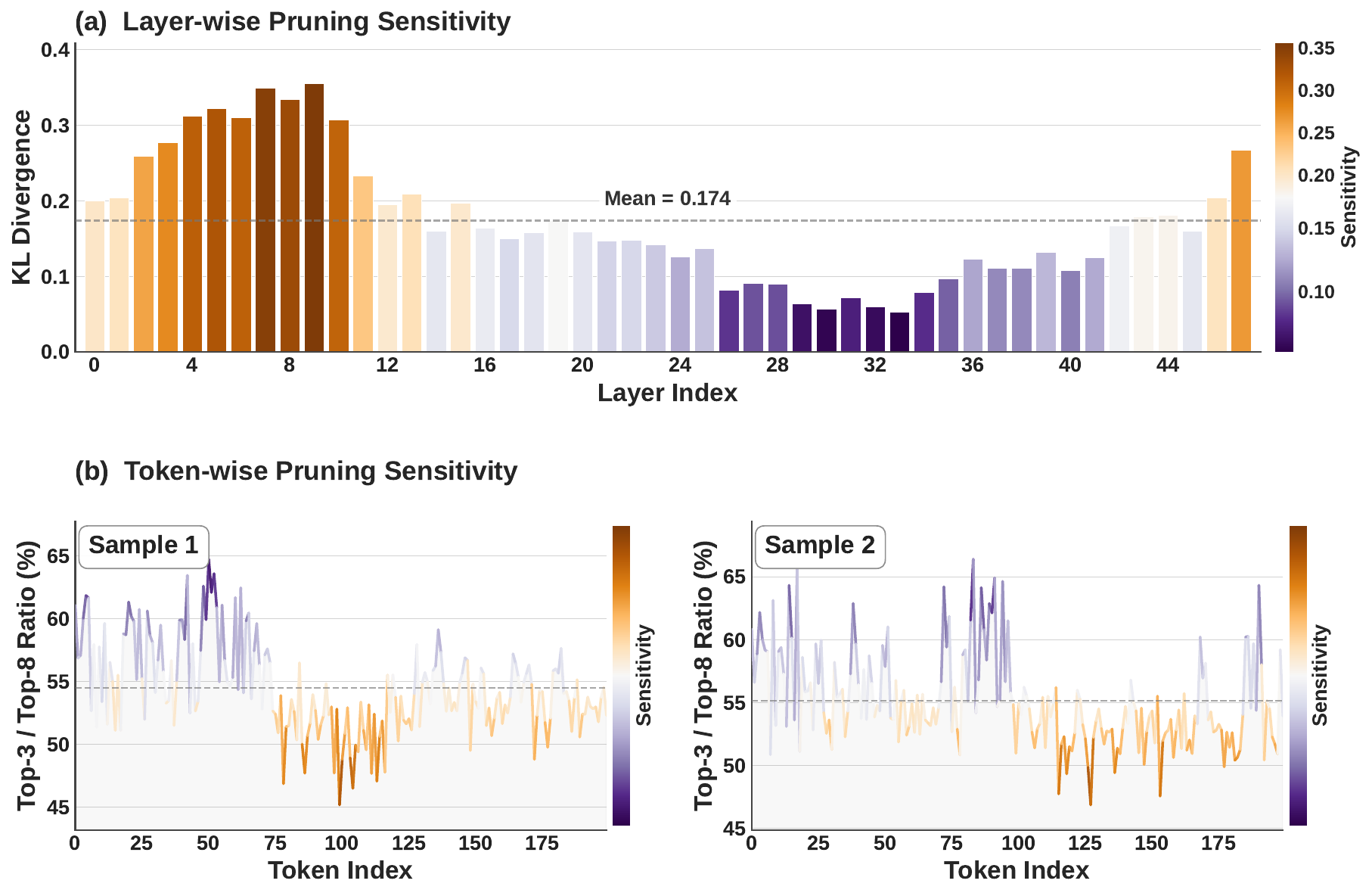}
    \caption{Sensitivity analysis of Qwen3-30B-A3B:
    (a) \textbf{layer-wise} pruning sensitivity measured by KL divergence across layers;
    (b) \textbf{token-wise} pruning sensitivity measured by the Top-3 / Top-8 routing weights ratio on two random samples.
    Both dimensions exhibit large variance, motivating a dynamic pruning strategy.}
    \label{fig:sensitivity_analysis}
    \vspace{-10pt}
\end{figure}

\paragraph{Layer Sensitivity.}  
We measure layer-wise pruning sensitivity by reducing the number of selected experts from $k=8$ to $k=3$ in each layer, and computing the Top‑1000 KL divergence between the output logits of the original and pruned models on MathQA calibration set. The results (Figure~\ref{fig:sensitivity_analysis}a) reveal substantial differences across layers: while some layers, especially at the begin and end of the model, are highly sensitive to pruning (shown in orange), many middle layers are much more robust (shown in purple). This large variance highlights the necessity of treating different layers differently when designing pruning strategies.

\paragraph{Token Sensitivity.}  
We further analyze token-level pruning sensitivity in long reasoning processes by examining how routing weights are distributed across experts. To quantify this, we compute the ratio between the cumulative routing weights of the top-3 and top-8 chosen experts, averaged across all layers. Figure~\ref{fig:sensitivity_analysis}b plots these ratios for the first 200 generated tokens in reasoning processes from two randomly selected GPQA-Diamond samples. The both results reveal substantial variability across tokens: some tokens concentrate most of their weights on a few experts and are thus robust to expert pruning, whereas others distribute weights evenly and are more sensitive.

\paragraph{Dynamic pruning.}  
Before combining layer‑level and token‑level information, we normalize their sensitivity scores to the range $[0,1]$.
Here, $W_l$ denotes the KL divergence of layer $l$ when reducing expert selection from $k=8$ to $k=3$.
$W_{\min}$ and $W_{\max}$ are defined as the divergences of the least‑sensitive and most‑sensitive layers, respectively.
For tokens, $R_i$ represents the top‑3/top‑8 routing weight ratio of token $i$ at a specific layer during inference, with $R_{\min}$ and $R_{\max}$ similarly estimated from the calibration set.
The normalized sensitivity scores are then defined as:
\[
\begin{aligned}
L'_l &= (W_l - W_{\min}) / (W_{\max} - W_{\min}), \\
T'_i &= (R_{\max} - R_i) / (R_{\max} - R_{\min}).
\end{aligned}
\]

The combined sensitivity score for token $i$ at layer $l$ is then:
\[
S_{i,l} = \lambda \cdot \tfrac{1}{2}\big(L'_l + T'_i\big), \quad \lambda \in (0,1)
\]
where we simply average layer-level and token-level sensitivities to balance their contributions, and use $\lambda$ to control the aggressiveness of pruning.  
The dynamic number of selected experts is determined as:
\[
K_{i,l} = \left\lfloor K_{\min} + (K_{\text{base}} - K_{\min}) \cdot S_{i,l} \right\rceil .
\]
Here $K_{\text{base}}=8$ is the default number of experts, and $K_{\min}=3$ serves as a safeguard to prevent selecting too few experts: from empirical inspection, we found that reducing the number of selected experts below this threshold leads to a complete collapse of model. 
We provide detailed hyperparameter sensitivity analysis in Appendix~\ref{app:pruning_param}, demonstrating that $K_{\min}$ and $\lambda$ are robust and easy to set in practice.

\begin{table*}[thp]
    \small
    \centering
    \caption{Comparison of \textbf{Pick} against three baselines across four models. Reported accuracies cover multiple benchmarks, with GSM8K \citep{cobbe2021gsm8k} only for DeepSeek-V2-Lite-Chat. (more results in Appendix~\ref{app:more_results})}
    \label{tab:pick}
    \resizebox{\textwidth}{!}{
    \begin{tabular}{|c|c|c|c|c|c|c|c|}
    \hline
       Models & Methods & GSM8K/AIME24 & Math500 & GPQA-Diamond & MMLU-Pro & LiveCodeBench & HumanEval+ \\ 
    \hline
    \multirow{4}{*}{\makecell{\includegraphics[height=1.2em]{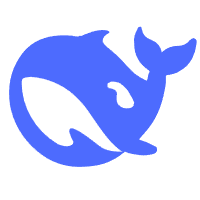} \\ Deepseek-V2-Lite-Chat}} 
                          & Baseline & 66.34 & 26.40 & 24.75 & 27.44 & 9.25 & 50.61 \\ \cline{2-8}
                          & Dynamic  & 66.56 & 26.80 & 24.75 & 27.03 & 9.00 & 50.61 \\
                          & Tips     & 66.87 & 26.60 & 25.25 & 28.15 & 9.50 & 51.22 \\
                          & \cellcolor{blue!7}Pick & \cellcolor{blue!7}\textbf{67.48} & \cellcolor{blue!7}\textbf{28.00} & \cellcolor{blue!7}\textbf{25.76} & \cellcolor{blue!7}\textbf{28.97} & \cellcolor{blue!7}\textbf{10.50} & \cellcolor{blue!7}\textbf{51.83} \\ \hline
    \multirow{4}{*}{\makecell{\includegraphics[height=1.2em]{deepseek.png} \\ Deepseek-v2.5-1210}} 
                          & Baseline & 23.33 & 81.00 & 47.47 & 27.97 & 60.75 & 83.54 \\ \cline{2-8}
                          & Dynamic  & 24.00 & 81.20 & 46.97 & 28.86 & 61.00 & 83.54 \\
                          & Tips     & 24.67 & 80.60 & 47.98 & 28.54 & 61.25 & 84.15 \\
                          & \cellcolor{blue!7}Pick & \cellcolor{blue!7}\textbf{27.34} & \cellcolor{blue!7}\textbf{82.80} & \cellcolor{blue!7}\textbf{49.49} & \cellcolor{blue!7}\textbf{30.06} & \cellcolor{blue!7}\textbf{62.75} & \cellcolor{blue!7}\textbf{84.76} \\ \hline
    \multirow{4}{*}{\makecell{\includegraphics[height=1.2em]{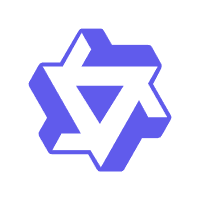} \\ Qwen3-30B-A3B}} 
                          & Baseline & 80.67 & 95.20 & 65.66 & 63.95 & 91.75 & 71.34 \\ \cline{2-8}
                          & Dynamic  & 82.66 & 94.80 & 66.16 & 63.41 & 92.00 & 71.95 \\
                          & Rice     & 83.33 & 95.60 & 67.68 & 64.02 & 93.50 & 72.56 \\
                          & \cellcolor{blue!7}Pick & \cellcolor{blue!7}\textbf{85.33} & \cellcolor{blue!7}\textbf{96.60} & \cellcolor{blue!7}\textbf{69.19} & \cellcolor{blue!7}\textbf{65.34} & \cellcolor{blue!7}\textbf{94.25} & \cellcolor{blue!7}\textbf{75.61} \\ \hline
    \multirow{4}{*}{\makecell{\includegraphics[height=1.2em]{qwen.png} \\ Qwen3-235B-A22B}} 
                          & Baseline & 84.67 & 96.00 & 71.21 & 68.45 & 92.75 & 78.05 \\ \cline{2-8}
                          & Dynamic  & 85.33 & 96.00 & 71.72 & 69.02 & 93.00 & 78.66 \\
                          & Rice     & 86.67 & 96.20 & 72.22 & 68.75 & 93.00 & 79.88 \\
                          & \cellcolor{blue!7}Pick & \cellcolor{blue!7}\textbf{88.00} & \cellcolor{blue!7}\textbf{96.80} & \cellcolor{blue!7}\textbf{74.24} & \cellcolor{blue!7}\textbf{70.18} & \cellcolor{blue!7}\textbf{94.50} & \cellcolor{blue!7}\textbf{81.10} \\ \hline
    \end{tabular}}
    \vspace{-10pt}
\end{table*}

\section{Experiments}
In this section, we first introduce the experimental setup, including models, datasets, and implementation details. We then separately compare our proposed \textbf{Pick} and \textbf{Ban} approaches with existing methods, highlighting their respective strengths. Finally, we show how combining \textbf{Ban\&Pick} achieves both improved performance and faster inference. 

\subsection{Experiment Setup}
\paragraph{Models and Datasets}
We conduct experiments on four fine-grained MoE models from two different model families, covering a wide range of parameter scales: DeepSeek-V2-Lite-Chat (16B), DeepSeek-V2.5-1210 (236B), Qwen3-30B-A3B, and Qwen3-235B-A22B. Our evaluation spans three categories of tasks using five commonly adopted datasets: AIME2024 (with 5 runs) \citep{patel2024aimeaioptimizationmultiple} and Math-500 for mathematical reasoning, GPQA-Diamond \citep{rein2023gpqagraduatelevelgoogleproofqa} and MMLU-Pro~\citep{wang2024mmluprorobustchallengingmultitask} for general knowledge reasoning, and LiveCodeBench \citep{jain2024livecodebench} together with HumanEval+ \citep{evalplus} for code generation. For DeepSeek-V2-Lite-Chat, we replace AIME2024 with GSM8K, as AIME2024 is overly difficult for this model. 

\paragraph{Implementation Details}
For the Pick method, we follow the procedures described in 
~\Cref{sec:pick} to identify key experts for each of the three task categories--math, code and general reasoning--and apply targeted enhancements to them. We then evaluate the resulting models on their corresponding task types. 
For the Ban method, we set $K_{\min}=3$ and $\lambda=0.7$ for all four models, aiming to achieve a balanced trade-off between accuracy and speedup (further discussed in Appendix~\ref{app:pruning_param}). When combining Pick and Ban, we note that in layers containing key experts, Ban has already pruned redundant experts, so directly applying the range-based replacement (strategy D) leads to noticeable performance degradation. To address this, we instead adopt the range-based addition (strategy C).

\begin{table*}[thp]
    \normalsize
    \centering
    \caption{Comparison of \textbf{Ban} against DES and ODP across four models. We report average experts per token (Avg-Topk), accuracy, and inference speedup under the \texttt{vLLM} (0.9.1) backend.}
    \label{tab:ban}
    \resizebox{\textwidth}{!}{
    \begin{tabular}{|c|c|c|c|c|c|c|c|c|c|}
    \hline
       Models & Methods & Avg-Topk & GSM8K/AIME24 & Math500 & GPQA-Diamond & MMLU-Pro & LiveCodeBench & HumanEval+ & Speedup \\ 
    \hline
    \multirow{4}{*}{\makecell{\includegraphics[height=1.2em]{deepseek.png} \\ Deepseek-V2-Lite-Chat}} 
                          & Baseline & 6.00 & 66.34 & 26.40 & 24.75 & 27.44 & 9.25 & 50.61 & 1.00 \\ \cline{2-10}
                          & DES      & 4.18 & 63.76 & 21.40 & 22.22 & 24.33 & 7.25 & 45.73 & 1.20 \\
                          & ODP      & 4.22 & 64.76 & 23.20 & 23.23 & 25.67 & 7.75 & 46.95 & 1.19 \\
                          & \cellcolor{blue!5}Ban & \cellcolor{blue!5}4.02 & \cellcolor{blue!5}\textbf{66.11} & \cellcolor{blue!5}\textbf{25.00} & \cellcolor{blue!5}\textbf{24.24} & \cellcolor{blue!5}\textbf{26.51} & \cellcolor{blue!5}\textbf{8.50} & \cellcolor{blue!5}\textbf{48.17} & \cellcolor{blue!5}\textbf{1.22} \\ \hline
    \multirow{4}{*}{\makecell{\includegraphics[height=1.2em]{deepseek.png} \\ Deepseek-v2.5-1210}} 
                          & Baseline & 6.00 & 23.33 & 81.00 & 47.47 & 27.97 & 60.75 & 83.54 & 1.00 \\ \cline{2-10}
                          & DES      & 4.21 & 22.66 & 80.20 & 39.90 & 23.89 & 58.00 & 79.27 & 1.21 \\
                          & ODP      & 4.29 & 23.33 & 80.20 & 43.43 & 24.42 & 59.25 & 81.71 & 1.20 \\
                          & \cellcolor{blue!5}Ban & \cellcolor{blue!5}3.97 & \cellcolor{blue!5}\textbf{23.33} & \cellcolor{blue!5}\textbf{80.60} & \cellcolor{blue!5}\textbf{45.96} & \cellcolor{blue!5}\textbf{26.85} & \cellcolor{blue!5}\textbf{60.25} & \cellcolor{blue!5}\textbf{82.93} & \cellcolor{blue!5}\textbf{1.24} \\ \hline
    \multirow{4}{*}{\makecell{\includegraphics[height=1.2em]{qwen.png} \\ Qwen3-30B-A3B}}
                          & Baseline & 8.00 & 80.67 & 95.20 & 65.66 & 63.95 & 91.75 & 71.34 & 1.00 \\ \cline{2-10}
                          & DES      & 4.94 & 73.33 & 92.60 & 56.06 & 60.12 & 84.50 & 65.24 & 1.25 \\
                          & ODP      & 4.98 & 74.67 & 93.20 & 60.10 & 62.31 & 86.00 & 67.07 & 1.24 \\
                          & \cellcolor{blue!5}Ban & \cellcolor{blue!5}4.82 & \cellcolor{blue!5}\textbf{80.00} & \cellcolor{blue!5}\textbf{94.60} & \cellcolor{blue!5}\textbf{64.65} & \cellcolor{blue!5}\textbf{63.18} & \cellcolor{blue!5}\textbf{90.00} & \cellcolor{blue!5}\textbf{70.12} & \cellcolor{blue!5}\textbf{1.25} \\ \hline
    \multirow{4}{*}{\makecell{\includegraphics[height=1.2em]{qwen.png} \\ Qwen3-235B-A22B}}
                          & Baseline & 8.00 & 84.67 & 96.00 & 71.21 & 68.45 & 92.75 & 78.05 & 1.00 \\ \cline{2-10}
                          & DES      & 4.89 & 68.00 & 94.80 & 62.12 & 63.87 & 90.75 & 70.12 & 1.26 \\
                          & ODP      & 4.91 & 73.33 & 95.20 & 66.16 & 66.04 & 91.00 & 73.78 & 1.26 \\
                          & \cellcolor{blue!5}Ban & \cellcolor{blue!5}4.77 & \cellcolor{blue!5}\textbf{83.33} & \cellcolor{blue!5}\textbf{95.40} & \cellcolor{blue!5}\textbf{69.70} & \cellcolor{blue!5}\textbf{67.32} & \cellcolor{blue!5}\textbf{91.25} & \cellcolor{blue!5}\textbf{75.61} & \cellcolor{blue!5}\textbf{1.27} \\ \hline
    \end{tabular}}
\end{table*}

\begin{table*}[tbhp]
    \normalsize
    \centering
    \caption{Results of combining \textbf{Ban\&Pick} for a balanced combination of both improvements on performance and efficiency across four models. We report average experts per token, accuracy, and inference speedup under vLLM.}
    \label{tab:banpick}
    \resizebox{\textwidth}{!}{
    \begin{tabular}{|c|c|c|c|c|c|c|c|c|c|}
    \hline
       Models & Methods & Avg-Topk & GSM8K/AIME24 & Math500 & GPQA-Diamond & MMLU-Pro & LiveCodeBench & HumanEval+ & Speedup \\ 
    \hline
    \multirow{2}{*}{\makecell{\includegraphics[height=0.9em]{deepseek.png} Deepseek-V2-Lite-Chat}} 
                          & Baseline   & 6.00 & 66.34 & 26.40 & 24.75 & 27.44 & 9.25 & 50.61 & 1.00 \\ \cline{2-10}
                          & \cellcolor{green!10}Ban\&Pick & \cellcolor{green!10}4.03 & \cellcolor{green!10}66.94 & \cellcolor{green!10}26.80 & \cellcolor{green!10}25.25 & \cellcolor{green!10}28.03 & \cellcolor{green!10}9.50 & \cellcolor{green!10}50.00 & \cellcolor{green!10}1.22 \\ \hline
    \multirow{2}{*}{\makecell{\includegraphics[height=0.9em]{deepseek.png} Deepseek-v2.5-1210}} 
                          & Baseline   & 6.00 & 23.33 & 81.00 & 47.47 & 27.97 & 60.75 & 83.54 & 1.00 \\ \cline{2-10}
                          & \cellcolor{green!10}Ban\&Pick & \cellcolor{green!10}3.99 & \cellcolor{green!10}26.00 & \cellcolor{green!10}81.60 & \cellcolor{green!10}47.98 & \cellcolor{green!10}29.13 & \cellcolor{green!10}61.25 & \cellcolor{green!10}84.15 & \cellcolor{green!10}1.23 \\ \hline
    \multirow{2}{*}{\makecell{\includegraphics[height=1.0em]{qwen.png} Qwen3-30B-A3B}} 
                          & Baseline   & 8.00 & 80.67 & 95.20 & 65.66 & 63.95 & 91.75 & 71.34 & 1.00 \\ \cline{2-10}
                          & \cellcolor{green!10}Ban\&Pick & \cellcolor{green!10}4.83 & \cellcolor{green!10}84.66 & \cellcolor{green!10}95.80 & \cellcolor{green!10}68.18 & \cellcolor{green!10}64.51 & \cellcolor{green!10}92.85 & \cellcolor{green!10}72.56 & \cellcolor{green!10}1.25 \\ \hline
    \multirow{2}{*}{\makecell{\includegraphics[height=1.0em]{qwen.png} Qwen3-235B-A22B}}
                          & Baseline   & 8.00 & 84.67 & 96.00 & 71.21 & 68.45 & 92.75 & 78.05 & 1.00 \\ \cline{2-10}
                          & \cellcolor{green!10}Ban\&Pick & \cellcolor{green!10}4.79 & \cellcolor{green!10}86.67 & \cellcolor{green!10}96.20 & \cellcolor{green!10}72.73 & \cellcolor{green!10}68.98 & \cellcolor{green!10}93.25 & \cellcolor{green!10}78.66 & \cellcolor{green!10}1.26 \\ \hline
    \end{tabular}}
\end{table*}

\subsection{Results of Pick}

We compare our proposed \textbf{Pick} method against three representative baselines: \\
(1) \textbf{Dynamic Routing.}  
\citep{huang-etal-2024-harder} adaptively allocates experts according to routing weights, assigning more experts to harder tokens and fewer to easier ones, realized by accumulating experts until a threshold is reached. We set thresholds $\{0.7, 0.8, 0.9\}$ and report the best result. \\
(2) \textbf{Tip.}  
\citep{wang2025thoughtsplaceunderthinkingo1like} penalizes tokens that typically indicate switching paths, thereby encouraging the model to continue deepening its current line of reasoning.\\
(3) \textbf{Rice.}  
\citep{wang2025expertsneedsteeringthinking} leverages explicit \texttt{<think>} tokens to identify cognitive experts that specialize in reasoning. During inference, the routing weights of these experts are multiplied by a constant factor to amplify their influence. 

As shown in Table~\ref{tab:pick}, \textbf{Pick} achieves consistent and notable improvements across four models on five datasets, outperforming all three baselines. The gains are especially pronounced on the Qwen3 series, which we attribute to their stronger expert specialization. Remarkably, on the large-scale Qwen3-235B-A22B, whose performance is already comparable to leading closed-source models, Pick further boosts accuracy by 3.33\%, 3.03\%, and 3.05\% on AIME2024, GPQA, and HumanEval+, respectively. These results highlight the potential of leveraging key experts to further improve MoE model performance. 
Meanwhile, on DeepSeek-v2.5-1210, Pick achieves an average improvement of 2.19\%, demonstrating strong generalization capability. 
Additionally, we report results on a broader range of task categories in Appendix~\ref{app:more_results}.

\subsection{Results of Ban}

We compare our proposed \textbf{Ban} method against two mainstream dynamic expert-pruning approaches:\\
(1) \textbf{DES.}  
\citep{lu-etal-2024-experts} (Dynamic Expert Skip) dynamically prunes low-weight experts 
by observing that,
for easy tokens, the contribution of lower-ranked experts is often negligible.
Concretely, it decides whether to drop the secondary expert based on the ratio between expert routing weights.\\
(2) \textbf{ODP.}  
\citep{huang2024mc} (Online Dynamic Pruning) extends DES by adding a key-token protection mechanism. Tokens that receive high attention scores are exempted from pruning. 

In terms of accuracy, as shown in Table~\ref{tab:ban}, both DES and ODP suffer from severe degradation on these challenging tasks, with drops exceeding 5\% on certain datasets. 
In contrast, Ban limits the accuracy loss to within 1.5\% on most models and datasets, except HumanEval+, which we found sensitive to pruning, and even within 1\% on more than half of the results.
Notably, on Qwen3-30B-A3B, Ban reduces accuracy on AIME2024 by only 0.67\%--equivalent to misanswering just one additional problem out of five evaluations--and incurs merely 1.01\% degradation on GPQA-Diamond.

In addition to accuracy, we record the average number of selected experts per token and the relative speedup—computed from the total inference time across all datasets—under the \texttt{vLLM} backend to comprehensively evaluate inference efficiency. \textbf{Ban} restricts the average number of selected experts to around 4 for DeepSeek models, leading to more than 1.2$\times$ actual speedup. On Qwen3 models, Ban keeps the average expert count below 5, achieving about 1.25$\times$ speedup. These results demonstrate the strong ability of Ban to preserve model performance while delivering substantial efficiency gains.

\subsection{Results of Ban\&Pick}

By combining \textbf{Ban} and \textbf{Pick}, our unified framework leverages smarter routing strategies to simultaneously deliver free accuracy gains and inference acceleration across models and datasets, demonstrating strong practicality. As shown in Table~\ref{tab:banpick}, on Qwen3-30B-A3B, Ban\&Pick achieves an average accuracy improvement of 1.67\% together with a 1.25$\times$ speedup, while on the larger-scale Qwen3-235B-A22B, it yields consistant accuracy gain and efficient speedup. As two plug-and-play modules, Ban and Pick share the common goal of utilizing smarter routing to release the potential of fine-grained MoE, while offering flexible choices: higher performance (\textbf{Pick}), faster inference (\textbf{Ban}), or a balanced combination of both improvements on performance and efficiency (\textbf{Ban\&Pick}).

\section{Conclusion}
In this study, we reveal that in fine-grained MoE models, relatively balanced routing strategy underutilizes a small set of key experts which have a disproportionately large impact on task performance, and fixed expert activation introduces substantial redundancy. By selectively increasing the utilization of key experts and dynamically pruning redundant ones, we release the potential of routing and provide a novel and effective framework Ban\&Pick that improves both accuracy and efficiency.




\begin{table*}[t]
\centering
\small
\caption{Enhanced key experts (formatted as (layer, expert)) for different domains across models.}
\label{tab:enhanced-experts}
\begin{tabularx}{\textwidth}{lXXX}
\toprule
\textbf{Model} & \textbf{Math Experts} & \textbf{General Experts} & \textbf{Code Experts} \\
\midrule
Qwen3-30B-A3B & (2,37), (9,18), (21,18) & (5,19), (7,114), (13,105) & (6,40), (18,40) \\
Qwen3-235B-A22B & (11,121), (31,121), (51,94) & (13,57), (18,91), (62,101) & (7,56), (62,48), (82,48) \\
DeepSeek-V2-Lite-Chat & (5,14), (9,56), (17,25) & (4,36), (7,31), (14,22) & (8,1), (12,28) \\
DeepSeek-v2.5-1210 & (6,58), (32,5) & (9,22), (16,66) & (1,129), (22,36), (35,13) \\
\bottomrule
\end{tabularx}
\vspace{-5pt}
\end{table*}

\begin{table*}[t]
\centering
\small
\setlength{\tabcolsep}{4pt}
\caption{Task accuracies (\%) on Qwen3-30B-A3B under different combinations of enhanced experts.
A \checkmark\ indicates that the corresponding domain experts are enhanced, while \(\times\) means not enhanced.}
\label{tab:interaction-results}
\begin{tabular}{cccccccc}
\toprule
\textbf{Math} & \textbf{General} & \textbf{Code} & \textbf{AIME2024} & \textbf{MATH-500} & \textbf{GPQA-Diamond} & \textbf{LiveCodeBench} & \textbf{HumanEval+} \\
\midrule
\(\times\) & \(\times\) & \(\times\) & 80.67 & 95.20 & 65.66 & 91.75 & 71.34 \\
\midrule
\checkmark & \(\times\) & \(\times\) & 85.33\,\change{↑4.66} & 96.60\,\change{↑1.40} & 65.15\,\change{↓0.51} & 91.50\,\change{↓0.25} & 71.95\,\change{↑0.61} \\
\(\times\) & \checkmark & \(\times\) & 82.00\,\change{↑1.33} & 95.20\,\change{↑0.00} & 69.19\,\change{↑3.53} & 91.75\,\change{↑0.00} & 72.56\,\change{↑1.22} \\
\(\times\) & \(\times\) & \checkmark & 81.33\,\change{↑0.66} & 95.00\,\change{↓0.20} & 65.66\,\change{↑0.00} & 94.25\,\change{↑2.50} & 75.61\,\change{↑4.27} \\
\checkmark & \checkmark & \(\times\) & 84.67\,\change{↑4.00} & 96.20\,\change{↑1.00} & 67.68\,\change{↑2.02} & 92.00\,\change{↑0.25} & 71.34\,\change{↑0.00} \\
\checkmark & \(\times\) & \checkmark & 84.67\,\change{↑4.00} & 96.00\,\change{↑0.80} & 66.16\,\change{↑0.50} & 93.50\,\change{↑1.75} & 74.39\,\change{↑3.05} \\
\(\times\) & \checkmark & \checkmark & 80.67\,\change{↑0.00} & 94.80\,\change{↓0.40} & 68.18\,\change{↑2.52} & 92.75\,\change{↑1.00} & 75.00\,\change{↑3.66} \\
\checkmark & \checkmark & \checkmark & 84.00\,\change{↑3.33} & 96.00\,\change{↑0.80} & 67.68\,\change{↑2.02} & 93.00\,\change{↑1.25} & 74.39\,\change{↑3.05} \\
\bottomrule
\end{tabular}
\end{table*}

\section{Related Work}
\label{related_work}

\paragraph{Enhancing Expert Utilization in MoE.}
Most existing work attempts to improve expert utilization efficiency in MoE models by redesigning training objectives or expert architectures during pre-training \citep{yan2025tcmoe, huang-etal-2024-harder, jin2024moeacceleratingmixtureofexpertsmethods}. However, while these approaches are insightful, they often sacrifice training stability for flexibility. 
More recently, \citep{wang2025expertsneedsteeringthinking} proposes a post-training method specifically designed for MoE models that employ the \texttt{<think>} token, enhancing reasoning by amplifying the weights of Cognitive Experts.
\paragraph{Dynamic Expert Pruning of MoE.}
Dynamic expert pruning has emerged as a key technique for accelerating MoE inference. \citet{lu-etal-2024-experts} proposed a pruning strategy that simply skips experts with less weights: when the weight of the top‑1 expert exceeds that of the top‑2 expert by a given threshold, only the top‑1 expert is activated. 
Building on this, \citet{huang2024mc} introduced a critical token protection mechanism, ensuring that important tokens are less affected by prunin. 
However, the substantial sensitivity differences across layers of MoE models and across tokens in long chain‑of‑thought reasoning are often overlooked, leaving room for further optimization.

\section{Interaction Among Key Experts Across Tasks}
\label{across_domains}
In this section, we further investigate the scenario where all identified key experts across math, general reasoning, and code are enhanced simultaneously. We first summarize the enhanced experts for each task type in ~\Cref{tab:enhanced-experts}. Then, using Qwen3-30B-A3B as a case study, we report in Table~\ref{tab:interaction-results} the performance when enhancing key experts for a single domain, for domain pairs (Math+General, Math+Code, General+Code), and for all three domains together. This allows us to examine whether experts from different domains complement or interfere with each other when activated jointly, providing a more comprehensive understanding of cross-domain interactions.

From Table~\ref{tab:interaction-results}, we observe that enhancing multiple domains simultaneously still brings clear improvements. For instance, when both Math and General experts are enhanced, AIME2024 increases by +4.0\% and GPQA by +2.02\%. However, these gains are slightly smaller than enhancing each domain individually ($-0.66\%$ and $-1.51\%$, respectively). When all three domains are enhanced together, the model continues to outperform the baseline---+3.33\% on AIME2024, +2.02\% on GPQA, and +3.05\% on HumanEval+---though the gains are somewhat reduced compared to the best single-domain setting.
These results indicate that while domain-specific key experts do interfere to some extent, their combined enhancement remains clearly beneficial, 
confirming their crucial role.







\section*{Limitations}
\paragraph{Towards more rigorous theoretical and pre-training-level analysis.}  
In Section~\ref{sec:motivation}, we attribute the existence of under-utilized \emph{key experts} to the interplay between two widely observed properties of MoE pre-training: routers tend to converge substantially earlier than experts, while the auxiliary balancing loss caps the activation frequency of any single expert. We argue that this interplay causes a persistent mismatch between an expert's evolving competence and its largely frozen activation frequency, and we provide extensive empirical evidence confirming that such under-utilized yet highly impactful experts indeed exist across current fine-grained MoE models. Nevertheless, our analysis remains an inference-time observation rather than a rigorous theoretical proof. Due to limited computational resources, we are unable to reproduce the full pre-training process under controlled variations of routing dynamics or balancing objectives, which would be required to causally isolate the contribution of each factor. We leave two directions to future work: (i) a more formal theoretical characterization of how premature router convergence and load-balancing constraints jointly shape expert specialization, and (ii) addressing this issue directly at the pre-training stage, so that the full potential of fine-grained MoE can be unlocked at the source rather than only recovered at inference.

\paragraph{Room for more sophisticated key-expert enhancement strategies.} 
In Section~\ref{sec:enhancing}, after identifying and verifying the existence of key experts, we design four simple and straightforward enhancement strategies and select the final one through comparative experiments. On the one hand, our results demonstrate that merely increasing the activation frequency of the identified key experts is already sufficient to yield consistent performance gains on the corresponding task domains, which clearly confirms the latent potential of key experts in fine-grained MoE models. On the other hand, we acknowledge that the strategies explored in this work are deliberately simple and likely fall short of fully exploiting key experts. More fine-grained designs, such as input-adaptive enhancement, dynamic adjustment of routing scores, or jointly optimizing the selection and weighting of key experts, are likely to bring further improvements. We leave the exploration of such more sophisticated enhancement strategies, built on top of the framework proposed in this paper, to future work.

\section*{Ethical Considerations}
\label{sec:ethics}

This work focuses on identifying and enhancing key experts in fine-grained Mixture-of-Experts (MoE) large language models, which is an inference-time analysis and adjustment technique and does not involve the collection of new data or human subjects. All models and benchmarks used in our experiments are publicly available and employed in accordance with their respective licenses and intended research use. Since our method improves the task performance of fine-grained MoE models without requiring additional training or modification of model parameters, it may contribute positively by enabling more efficient utilization of existing pre-trained MoE models and reducing the need for costly retraining. We do not foresee any direct negative societal impact arising from this work beyond the general considerations applicable to large language models.


\section*{Acknowledgments}
This work was supported by the National Natural Science Foundation of China (Grant No. 62572471).


\bibliography{custom}

\appendix

\section{More Related Work}
\label{app:related_work}


\paragraph{Load Balance Loss and Expert Specialization.}
The tension between load balancing and expert specialization in MoE training has been increasingly recognized in recent literature. While the auxiliary load balance loss is widely adopted to prevent routing collapse and ensure even expert utilization, several works have observed that it inadvertently constrains expert specialization and degrades model performance \citep{wang2024auxiliary, guo2025advancingexpertspecializationbetter}. To mitigate this issue, prior work has explored two complementary directions: \citet{wang2024auxiliary} eliminate the auxiliary loss entirely and instead enforce balance through dynamically updated expert-wise biases on routing scores, while \citet{guo2025advancingexpertspecializationbetter} retain the auxiliary loss but introduce additional orthogonality and variance objectives to strengthen specialization. These findings collectively confirm that the load balance loss, despite its necessity for stable training, limits expert specialization—an observation that aligns with and supports the motivation of our work.

\paragraph{Enhancing Expert Utilization in MoE.}
Most existing work attempts to improve expert utilization efficiency in MoE models by redesigning training objectives or expert architectures during pre-training \citep{yan2025tcmoe, huang-etal-2024-harder, jin2024moeacceleratingmixtureofexpertsmethods}. 
TC-MoE \citep{yan2025tcmoe} expands the expert space by applying a ternary set $\{-1, 0, 1\}$ to each expert and introduces a tailored load-balance and reward loss, enabling more efficient activation patterns and a flexible trade-off between effectiveness and efficiency. 
\citet{huang-etal-2024-harder} break the fixed Top-$K$ paradigm by proposing a dynamic expert selection framework that adjusts the number of activated experts based on input difficulty via a confidence threshold, activating more experts for complex reasoning and fewer for simple inputs; they further observe that lower layers rely more heavily on expert combinations than upper layers. 
\citep{chen2026certainheaduncertaintail} introduces Expert-Sample, a training-free method that preserves high-confidence selections while injecting controlled stochasticity into the uncertain tail in fine-grained MoE, enabling diverse generation without destabilizing outputs.
MoE++ \citep{jin2024moeacceleratingmixtureofexpertsmethods} introduces a heterogeneous MoE framework that integrates standard FFN experts with three types of zero-computation experts (zero, copy, and constant), allowing each token to engage with a dynamic number of FFNs, be adjusted by constant vectors, or skip the MoE layer entirely, achieving $1.1{\sim}2.1\times$ throughput speedup over vanilla MoE. 
While these approaches are insightful, they often sacrifice training stability for flexibility, therefore hard to be applied to larger scale MoE.

More recently, \citet{wang2025expertsneedsteeringthinking} identify a small subset of experts that are highly correlated with the \texttt{<think>} token and play a disproportionate role in reasoning, which they term \emph{cognitive experts}; by amplifying the routing weights of these experts at inference time, they substantially enhance reasoning performance. 
However, this method is restricted to MoE models that explicitly employ the \texttt{<think>} token, limiting its general applicability.

\paragraph{Dynamic Expert Pruning of MoE.}
A growing body of work has observed substantial redundancy in the expert selection of MoE models, particularly in fine-grained MoE architectures where both the total number of experts and the number of activated experts per token are large \citep{lu-etal-2024-experts, huang2024mc, chen-etal-2025-eac, lasby2026reapexpertspruningprevails}. Motivated by this redundancy, dynamic expert pruning has emerged as a key technique for accelerating MoE inference. 
\citet{lu-etal-2024-experts} were the first to propose dynamic expert pruning, introducing a simple yet effective strategy that skips low-weight experts: when the routing weight of the top-1 expert exceeds that of the top-2 expert by a predefined threshold, only the top-1 expert is activated, thereby reducing computation without retraining. 
Building on this, \citet{huang2024mc} introduce a critical-token protection mechanism that exempts tokens with high attention scores from pruning, ensuring that performance-sensitive tokens retain their full set of experts. 
While the above approaches reduce computation from the \emph{token side} by adjusting the number of experts each token activates, \citet{chen-etal-2025-eac} approach the problem from the \emph{expert side}: rather than per-token decisions, they directly skip the inference of experts deemed globally less important, achieving more straightforward acceleration. 
More recently, \citet{lasby2026reapexpertspruningprevails} propose REAP (Router-weighted Expert Activation Pruning), a saliency criterion that jointly considers router gate values and expert activation norms, and computes a conditional average over the tokens an expert is activated on, thereby decoupling an expert's functional contribution from its raw activation frequency. 
Beyond the LLM setting, \citet{huang2026modesacceleratingmixtureofexpertsmultimodal} extend dynamic expert pruning to MoE-based multimodal LLMs, designing pruning strategies that exploit modality-specific differences and effectively accelerating MoE-MLLM inference. 

However, most of these works focus on non-reasoning tasks, and the substantial sensitivity differences across layers of MoE models and across tokens in long chain-of-thought reasoning are often overlooked, leaving room for further optimization.

\section{Details of Identifying Key Experts}
\label{app:identify-key-experts}
In Section~\ref{sec:pick-key-experts}, we introduced our method for identifying key experts by measuring the KL divergence between the output distributions of the original model and variants where individual domain-specific experts are pruned. There, we focused on Qwen3-30B-A3B with math tasks as a case study, to illustrate how only a small subset of high-frequency experts exert a decisive influence on model predictions. Here, we provide complete details of the identification procedure as well as comprehensive results across \textbf{four different models} and \textbf{three task types} (math, general, and code). For each model, we report the KL divergence values of candidate experts when pruned, visualized as bar charts.

\subsection{Selection Criterion of Key Experts}
\label{app:key-expert-criterion}
We first describe how the candidate set of key experts is filtered, which directly addresses the question of what specific divergence threshold is used to identify key experts. To prepare for this analysis, we collect the \textbf{domain-specialized experts} following the procedure in Section~\ref{sec:expert-specialization}. For each task type, we record the expert usage frequencies of the model on a calibration set. For the Qwen3 series, we define domain-specialized experts as those whose usage frequency exceeds six times the expected frequency under perfectly uniform routing. For the DeepSeek series, where the distribution of expert usage is less sharply peaked, we instead adopt a relaxed threshold of three times the uniform baseline. The experts identified in this way serve as the candidates for the subsequent KL divergence experiments.

Among these candidates, we then rank experts by their KL divergence values (computed as described in Section~\ref{app:kl-computation}) and select the top-ranked experts as the final key experts. Rather than fixing a single absolute KL threshold, we adopt a top-$N$ selection rule, where $N$ is determined per model and per domain by a clear gap in the sorted KL values: as illustrated in Figures~\ref{fig:kl-deepseek-v2}--\ref{fig:kl-qwen3-big}, the KL distribution exhibits a sharp drop after a small number of dominant experts, naturally separating the key experts from the rest. This avoids the brittleness of a single fixed numerical cutoff while still yielding a well-defined and reproducible selection.

\subsection{Calibration Sets for Each Task Type}
\label{app:calibration-sets}
This subsection specifies the calibration sets used for each of the three task types. We use MathQA for math, ARC-Challenge~\citep{allenai:arc} for general, and the college-computer-science-compilers subset of MMLU~\citep{hendryckstest2021} for code. We mainly select datasets where only the first token of the answer is required as output, because this simplifies evaluation while still providing an accurate measure of the KL divergence.

It is worth emphasizing two points regarding these calibration sets. First, none of them overlaps with the benchmarks used in our main experiments (e.g., GSM8K, AIME24, Math500, GPQA-Diamond, MMLU-Pro, LiveCodeBench, HumanEval+), so the identified key experts are not tuned on the evaluation data, ruling out concerns of data leakage. Second, the identification procedure is not sensitive to the specific calibration dataset chosen and does not overfit to it. The underlying reason is that fine-grained MoE models consistently exhibit similar expert selection frequencies across different datasets within the same task category~\citep{chen-etal-2025-eac}: math-related calibration sets activate a similar group of math-specialized experts, code-related calibration sets activate a similar group of code-specialized experts, and so on. As a result, key experts identified on one in-domain calibration set transfer well to other datasets of the same domain, which is also empirically validated by the strong cross-dataset generalization observed in our main results.

\subsection{KL Divergence Computation and Results}
\label{app:kl-computation}
\begin{figure*}[t]
    \centering
    \begin{minipage}{0.32\textwidth}
        \centering
        \includegraphics[width=\linewidth]{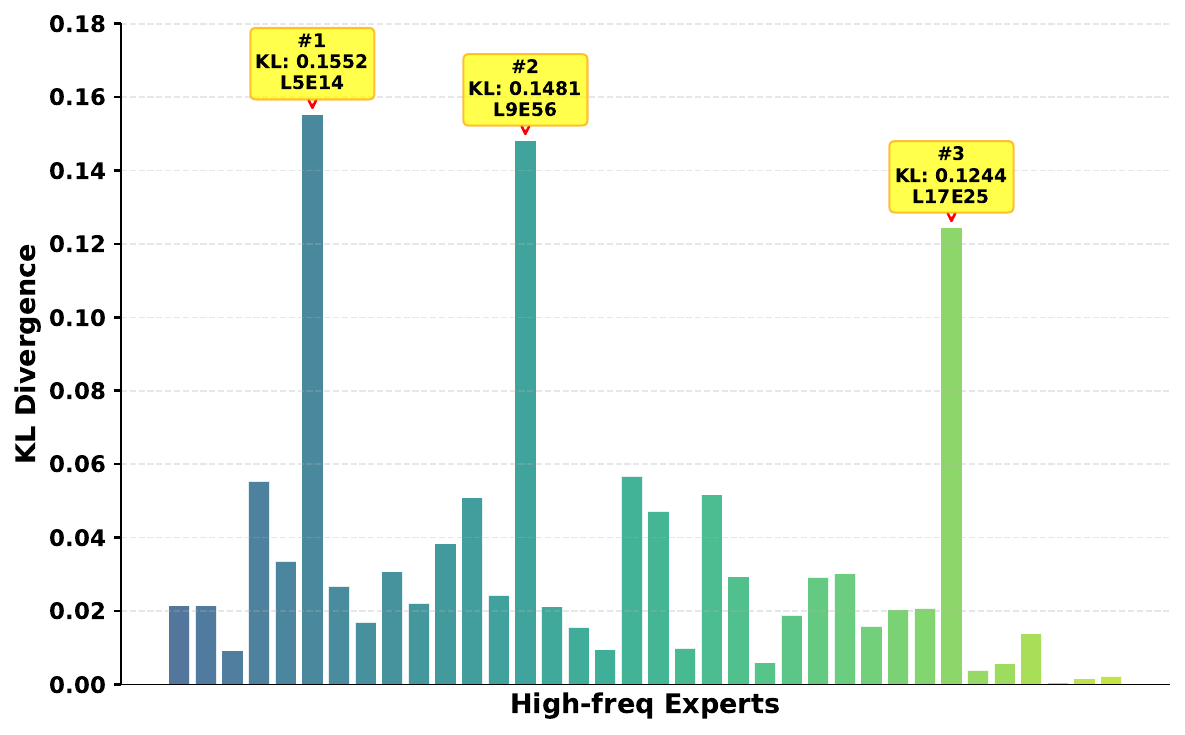}
        \centerline{(a) Math}
    \end{minipage}
    \begin{minipage}{0.32\textwidth}
        \centering
        \includegraphics[width=\linewidth]{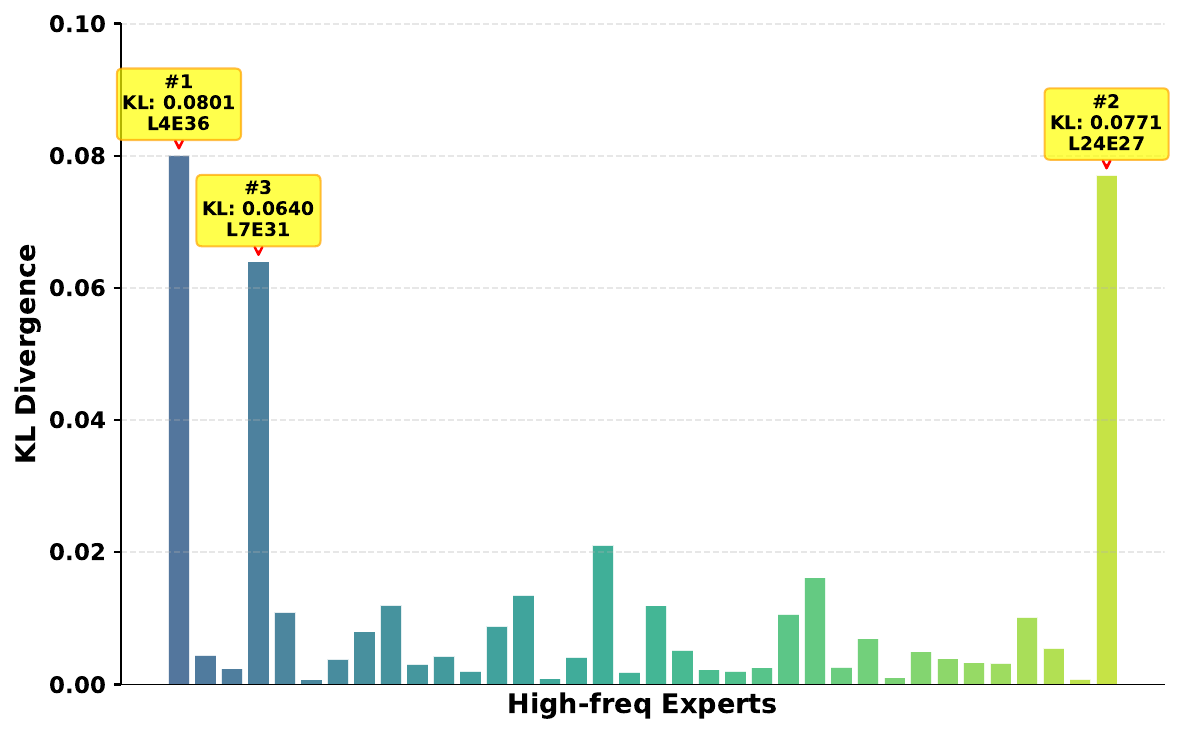}
        \centerline{(b) General}
    \end{minipage}
    \begin{minipage}{0.32\textwidth}
        \centering
        \includegraphics[width=\linewidth]{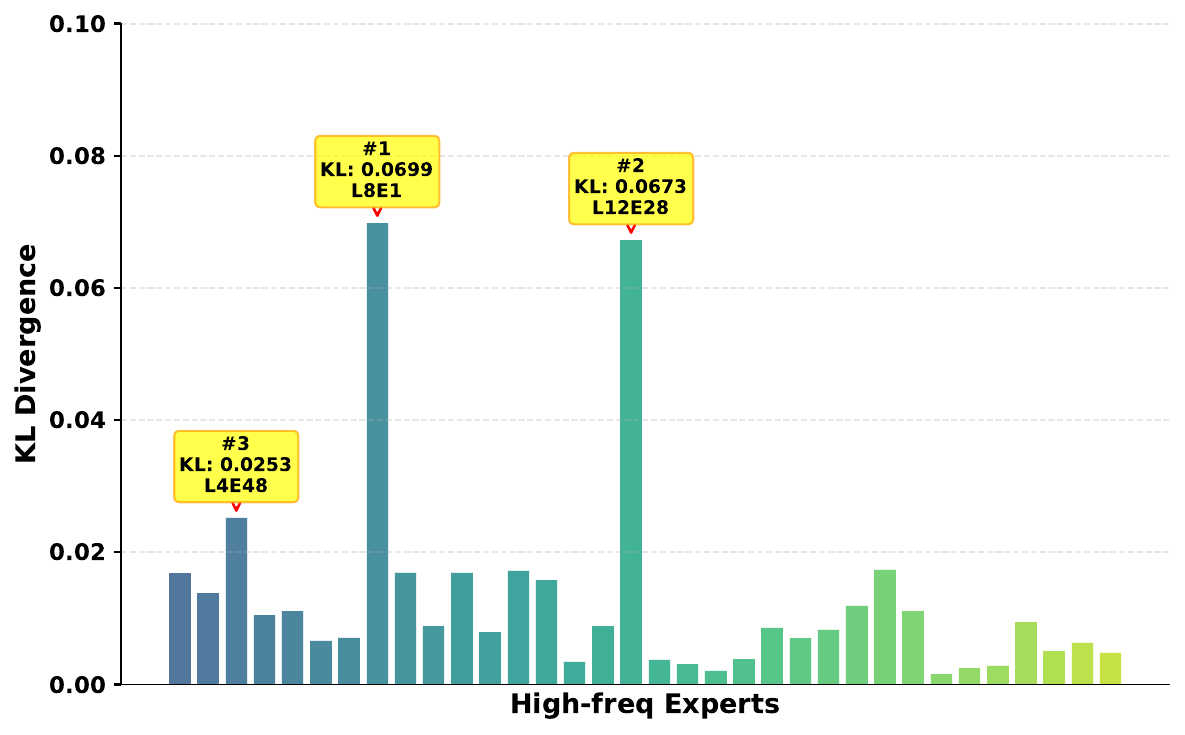}
        \centerline{(c) Code}
    \end{minipage}
    \caption{KL divergence of domain-specific experts when pruned, for \textbf{DeepSeek-V2-Lite-Chat} across math, general, and code tasks.}
    \label{fig:kl-deepseek-v2}
\end{figure*}

\begin{figure*}[t]
    \centering
    \begin{minipage}{0.32\textwidth}
        \centering
        \includegraphics[width=\linewidth]{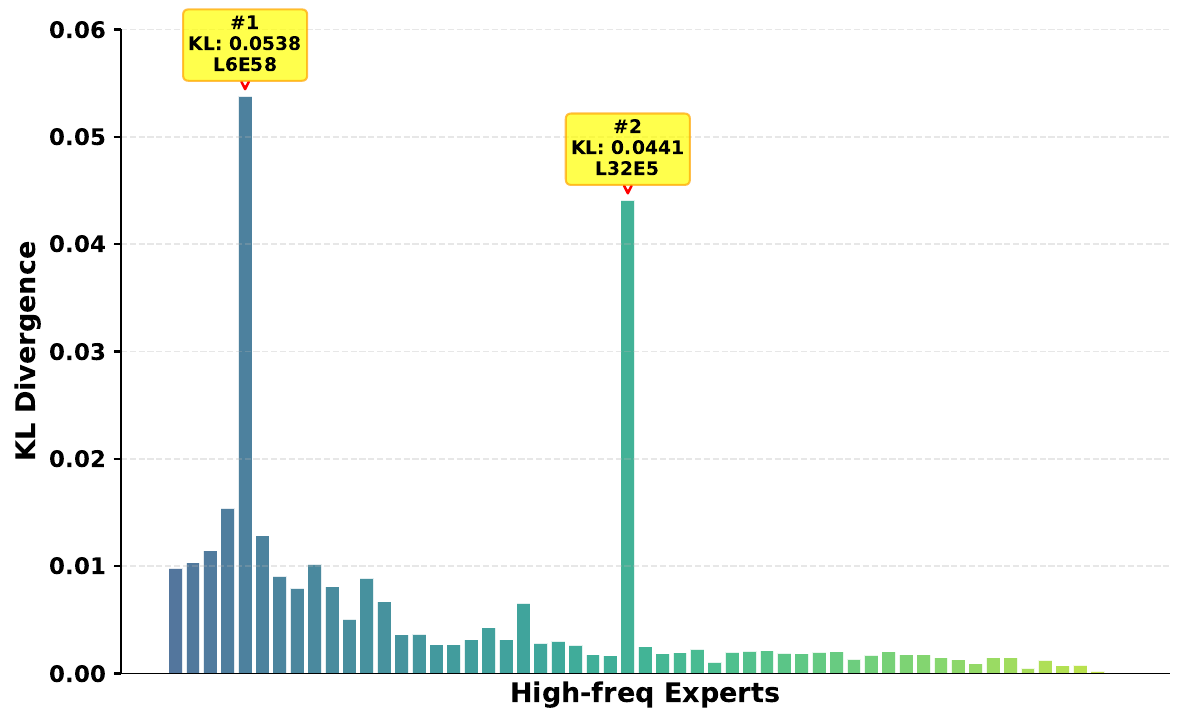}
        \centerline{(a) Math}
    \end{minipage}
    \begin{minipage}{0.32\textwidth}
        \centering
        \includegraphics[width=\linewidth]{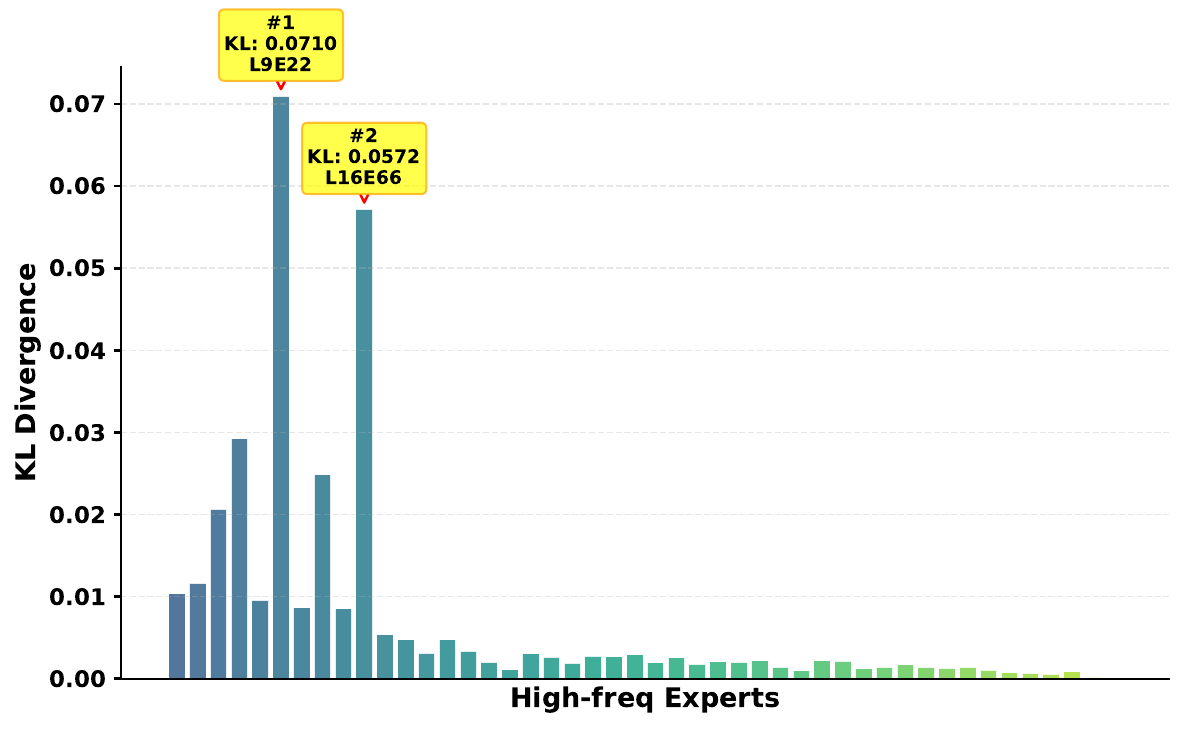}
        \centerline{(b) General}
    \end{minipage}
    \begin{minipage}{0.32\textwidth}
        \centering
        \includegraphics[width=\linewidth]{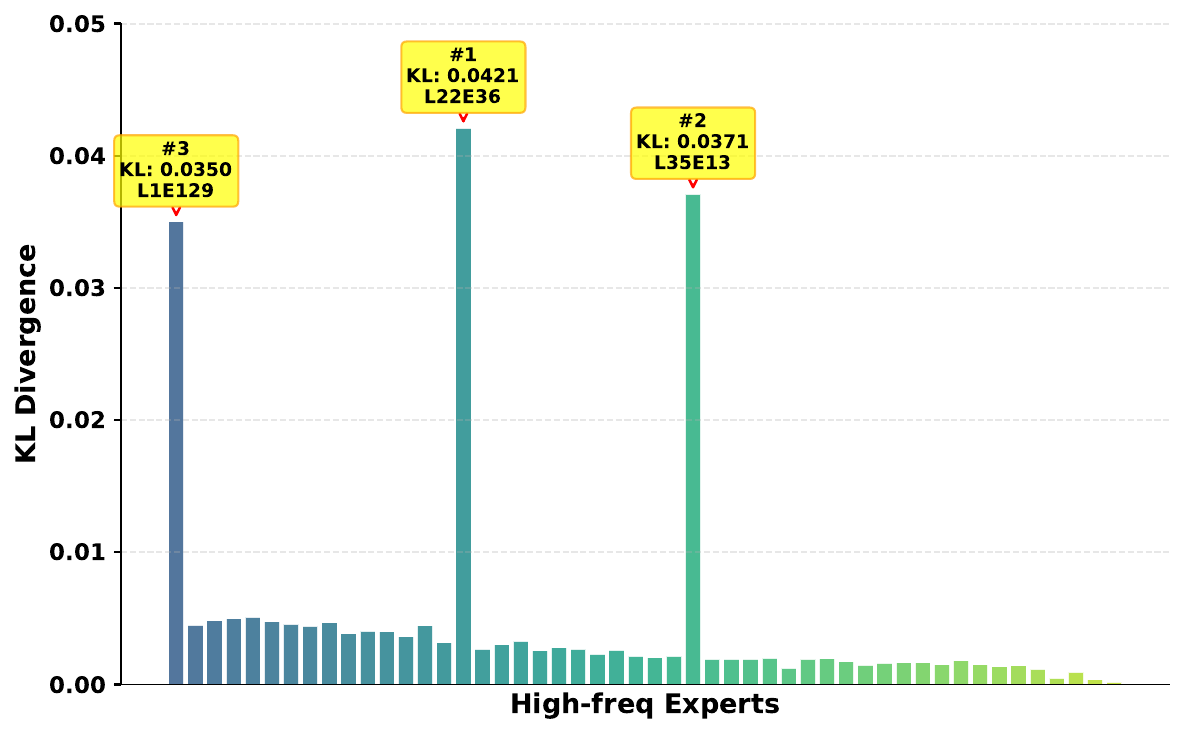}
        \centerline{(c) Code}
    \end{minipage}
    \caption{KL divergence of domain-specific experts when pruned, for \textbf{DeepSeek-v2.5-1210} across math, general, and code tasks.}
    \label{fig:kl-deepseek-v2.5}
\end{figure*}

\begin{figure*}[t]
    \centering
    \begin{minipage}{0.32\textwidth}
        \centering
        \includegraphics[width=\linewidth]{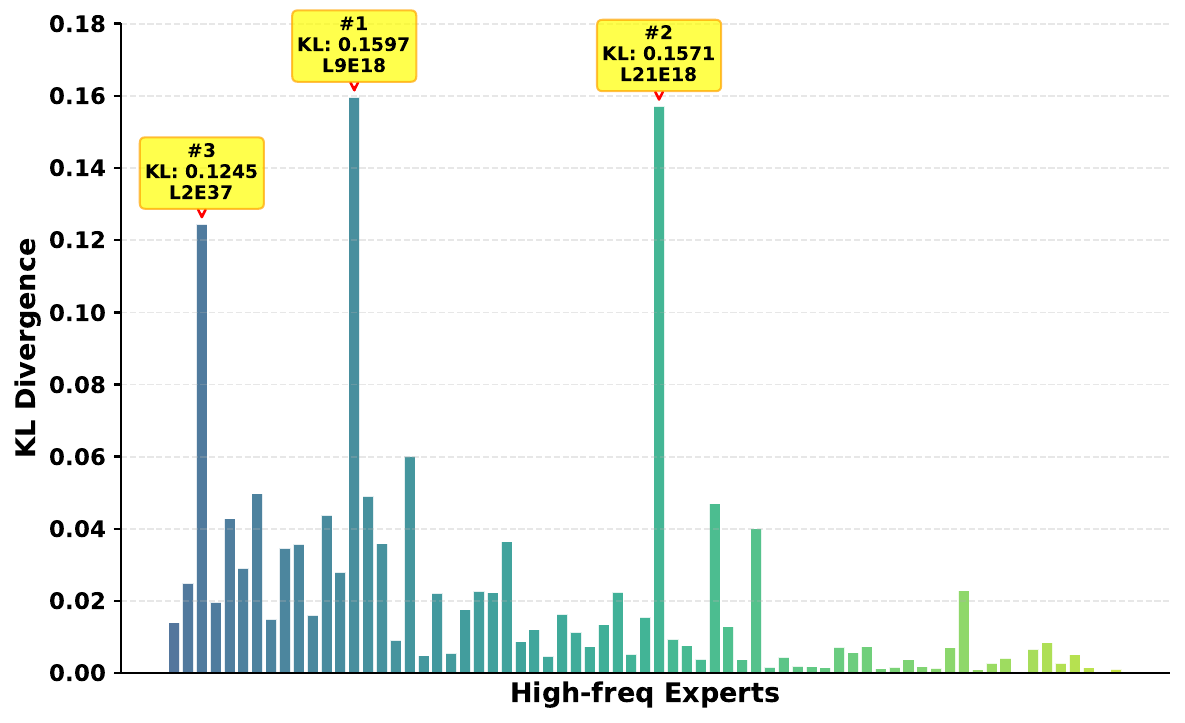}
        \centerline{(a) Math}
    \end{minipage}
    \begin{minipage}{0.32\textwidth}
        \centering
        \includegraphics[width=\linewidth]{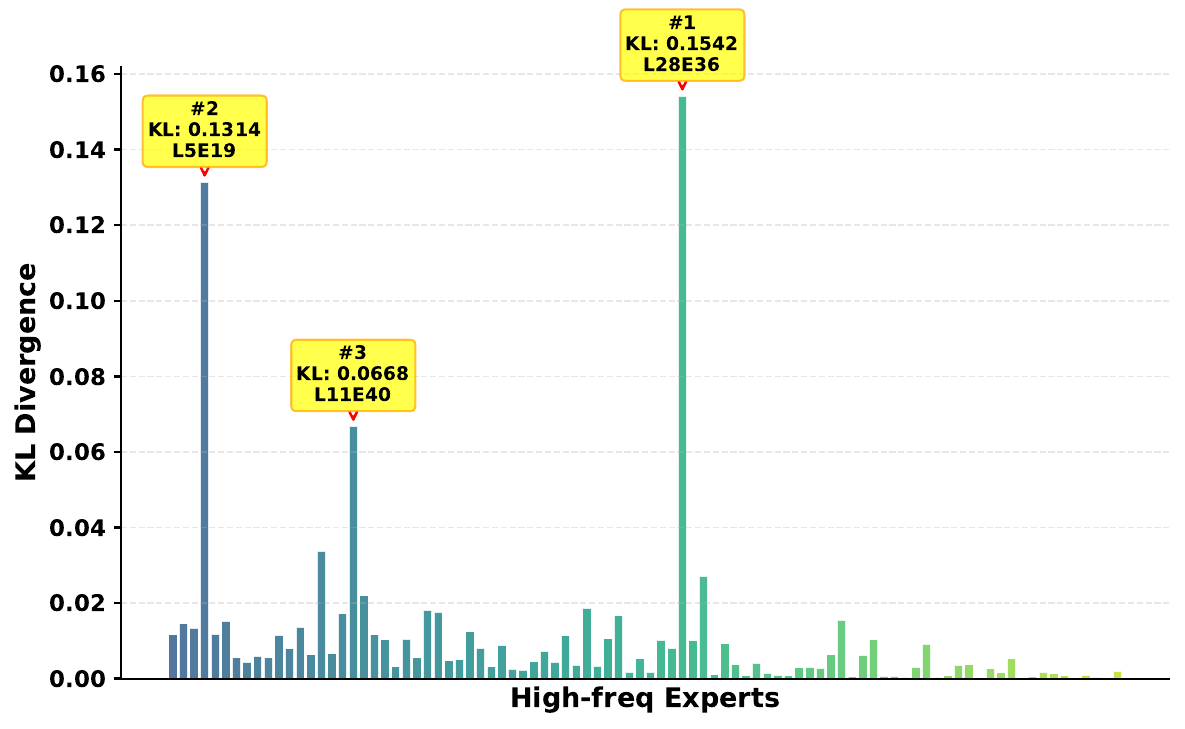}
        \centerline{(b) General}
    \end{minipage}
    \begin{minipage}{0.32\textwidth}
        \centering
        \includegraphics[width=\linewidth]{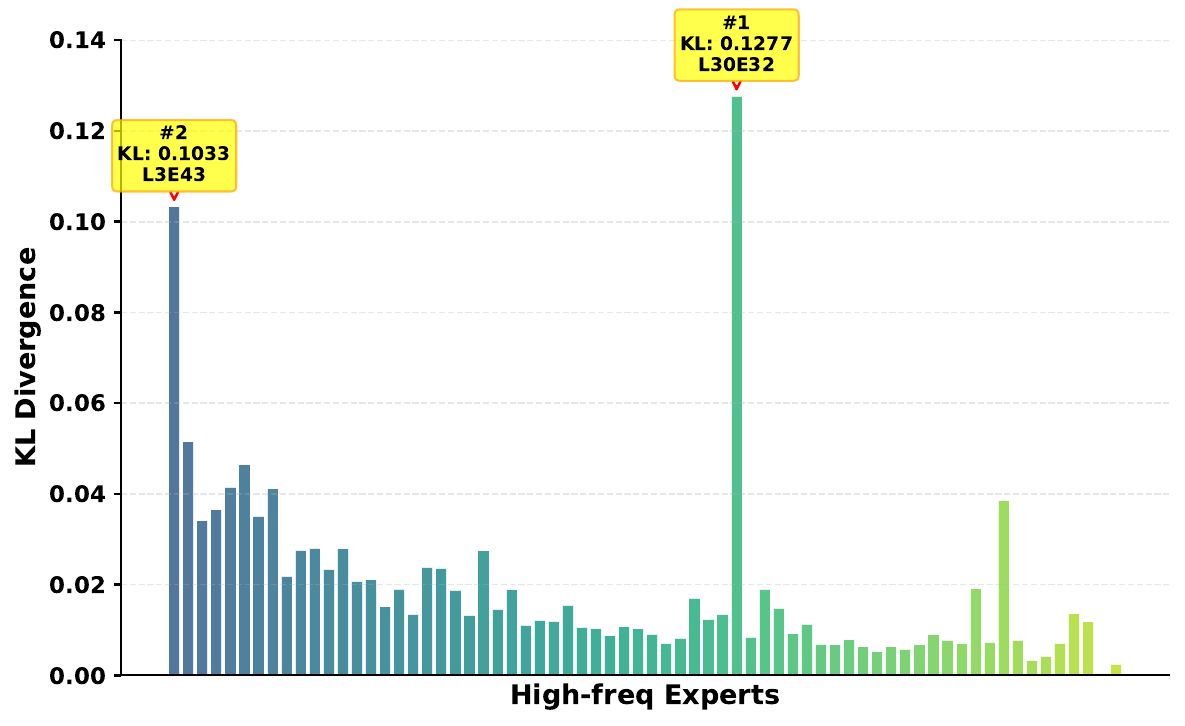}
        \centerline{(c) Code}
    \end{minipage}
    \caption{KL divergence of domain-specific experts when pruned, for \textbf{Qwen3-30B-A3B} across math, general, and code tasks.}
    \label{fig:kl-qwen3-30b}
\end{figure*}

\begin{figure*}[t]
    \centering
    \begin{minipage}{0.32\textwidth}
        \centering
        \includegraphics[width=\linewidth]{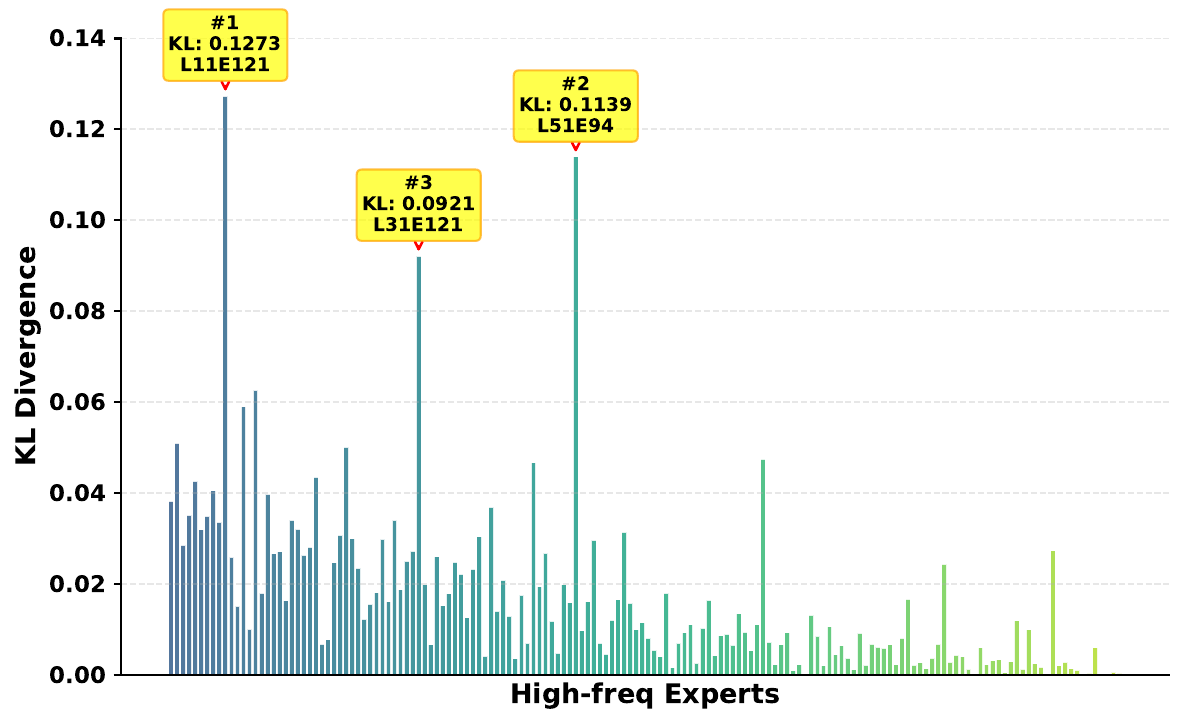}
        \centerline{(a) Math}
    \end{minipage}
    \begin{minipage}{0.32\textwidth}
        \centering
        \includegraphics[width=\linewidth]{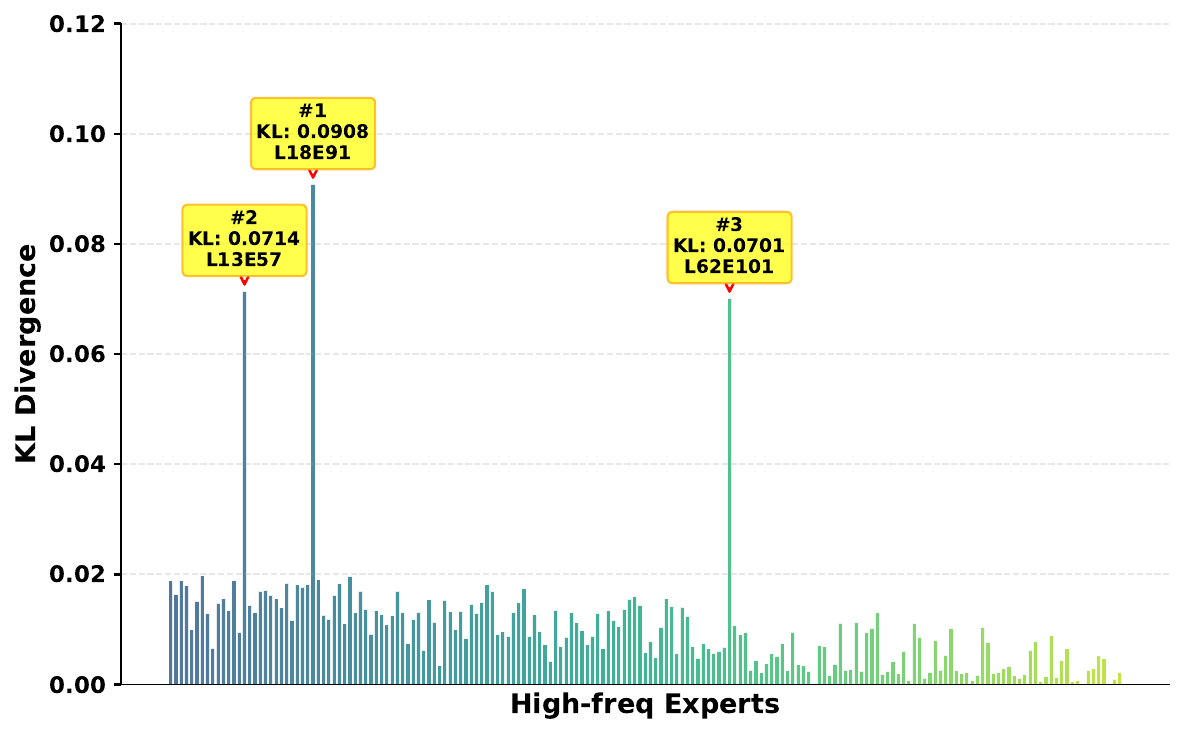}
        \centerline{(b) General}
    \end{minipage}
    \begin{minipage}{0.32\textwidth}
        \centering
        \includegraphics[width=\linewidth]{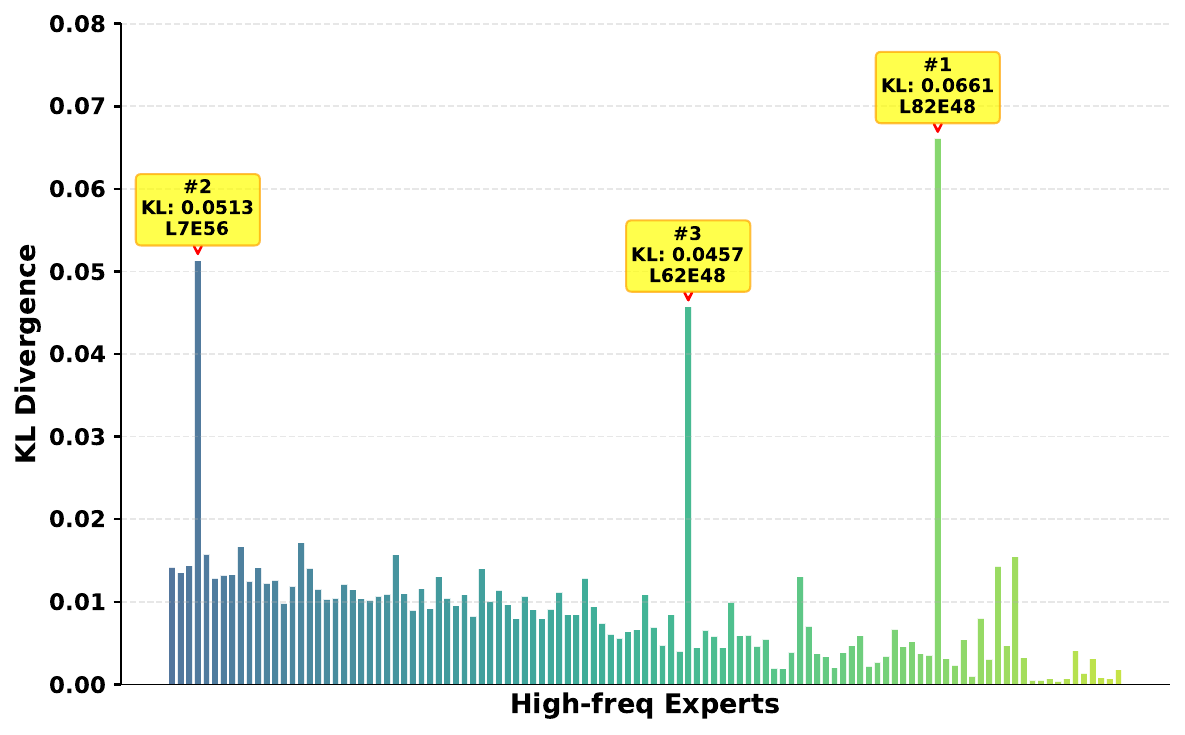}
        \centerline{(c) Code}
    \end{minipage}
    \caption{KL divergence of domain-specific experts when pruned, for \textbf{Qwen3-235B-A22B} across math, general, and code tasks.}
    \label{fig:kl-qwen3-big}
\end{figure*}

Following prior work~\citep{hu2025ostquant}, we note that the model's output logits typically follow a long-tail distribution, where only the top-ranked tokens carry substantial probability mass. Therefore, when computing KL divergence, we restrict attention to the \textbf{Top-1000 tokens}. Concretely, let \(p\) denote the probability distribution obtained from the original model, and \(q\) the distribution when a candidate expert is pruned. Denote by \(\mathcal{I}_{1000}\) the index set of the top-1000 tokens under \(p\). For each \(i \in \mathcal{I}_{1000}\), we define the restricted distributions as
\[
p'(i) = \frac{p(i)}{\sum_{j \in \mathcal{I}_{1000}} p(j)}.
\]
\[
q'(i) = \frac{q(i)}{\sum_{j \in \mathcal{I}_{1000}} q(j)}.
\]
The KL divergence used in our analysis is then
\[
D_{\mathrm{KL}}(p' \parallel q') = \sum_{i \in \mathcal{I}_{1000}} p'(i) \log \frac{p'(i)}{q'(i)}.
\]
This definition ensures that the comparison focuses on the most influential tokens, while ignoring the long tail of near-zero probabilities. The complete results for all models and task types are shown in Figures~\ref{fig:kl-deepseek-v2}--\ref{fig:kl-qwen3-big}. Consistent with the case study in the main text, we observe across different models and domains that only a small subset of high-frequency experts causes substantial divergence, while the majority of experts induce negligible changes.

\section{Complete Results on Expert Specialization}
\label{app:complete-result}
\begin{figure*}[t]
    \centering
    \includegraphics[height=0.95\textheight]{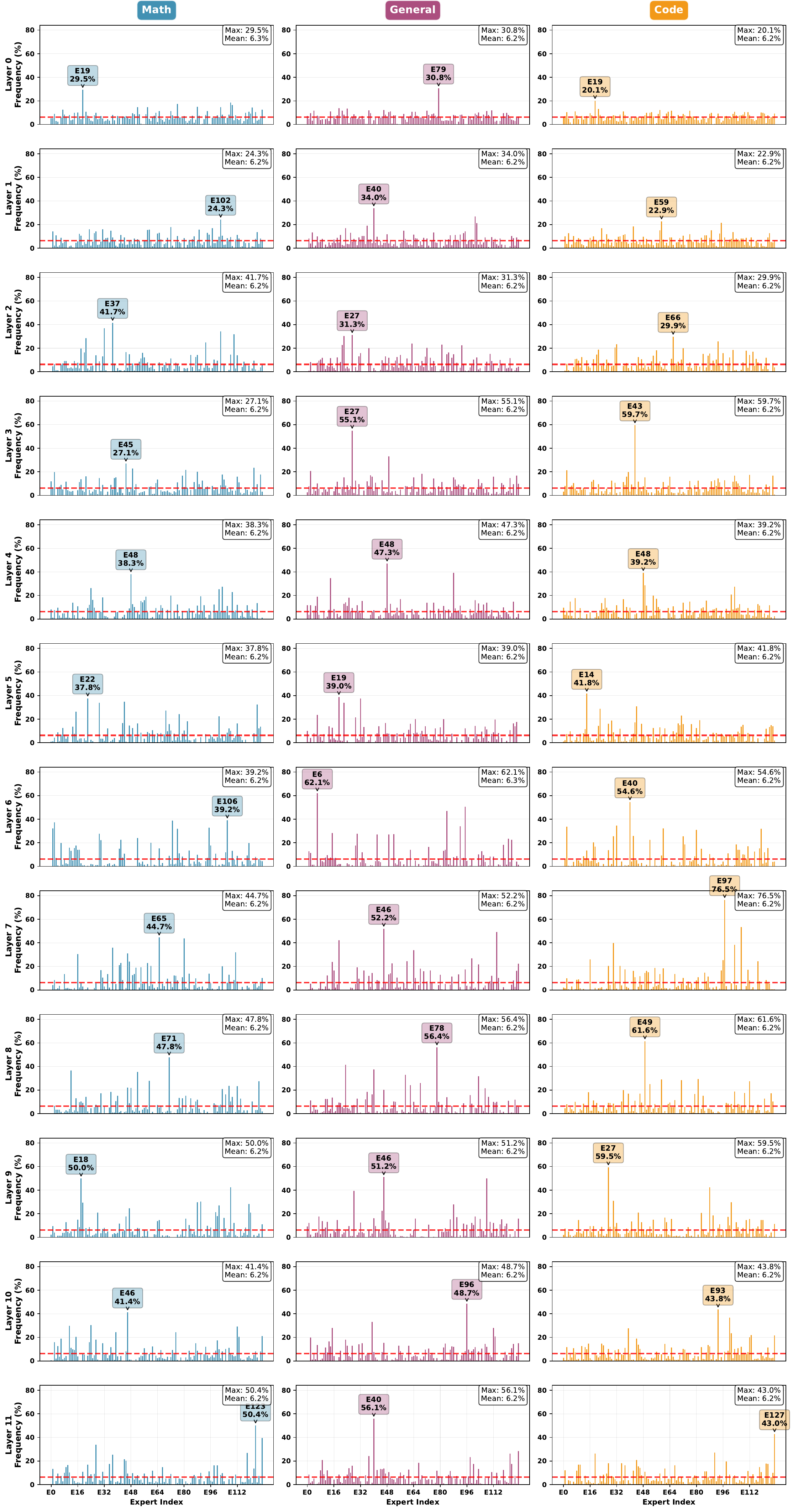}
    \caption{Expert usage frequency for layers 0--11 across math, code, and general tasks.}
    \label{fig:freq-0-11}
\end{figure*}

\begin{figure*}[t]
    \centering
    \includegraphics[height=0.95\textheight]{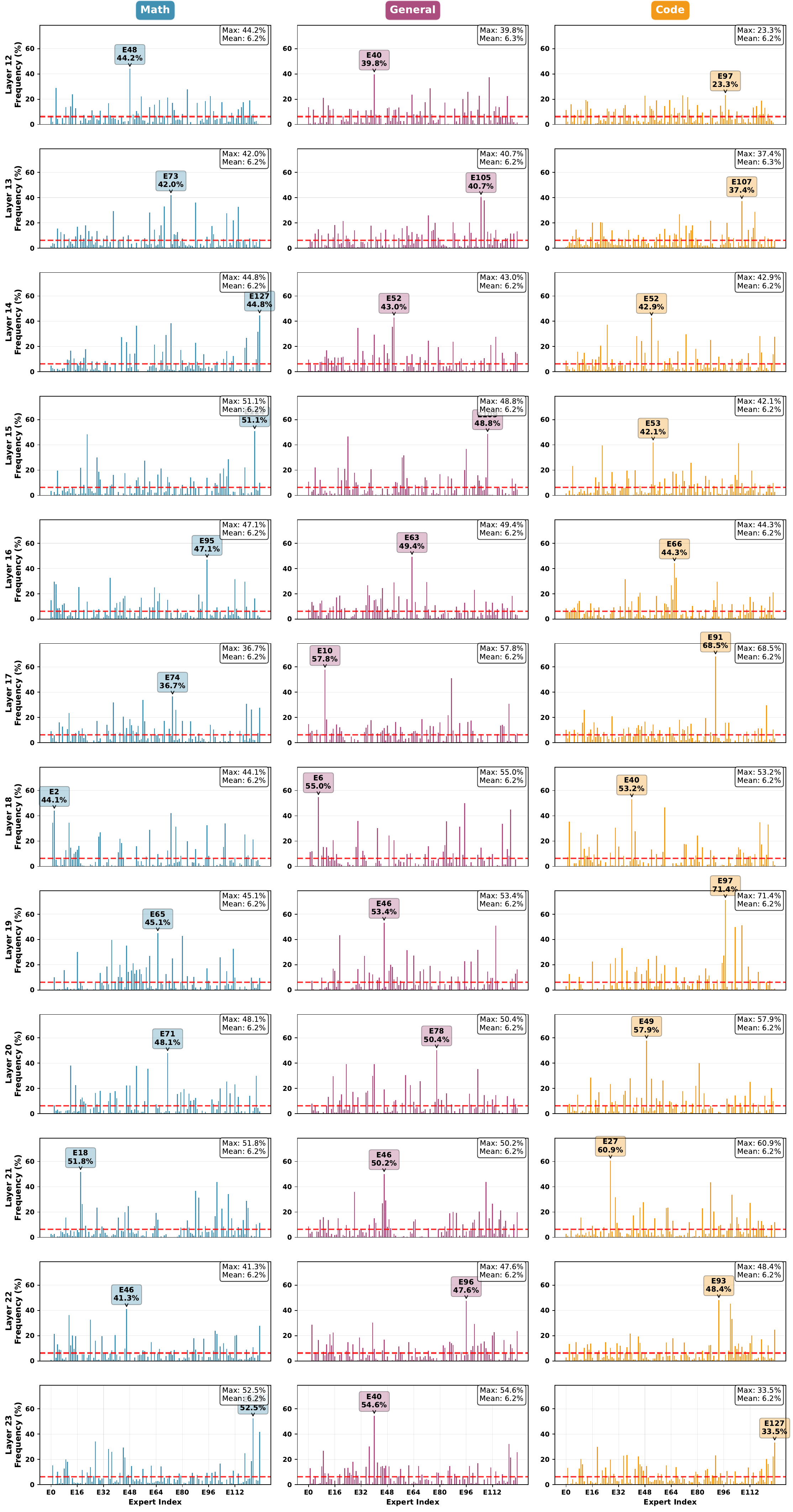}
    \caption{Expert usage frequency for layers 12--23 across math, code, and general tasks.}
    \label{fig:freq-12-23}
\end{figure*}

\begin{figure*}[t]
    \centering
    \includegraphics[height=0.95\textheight]{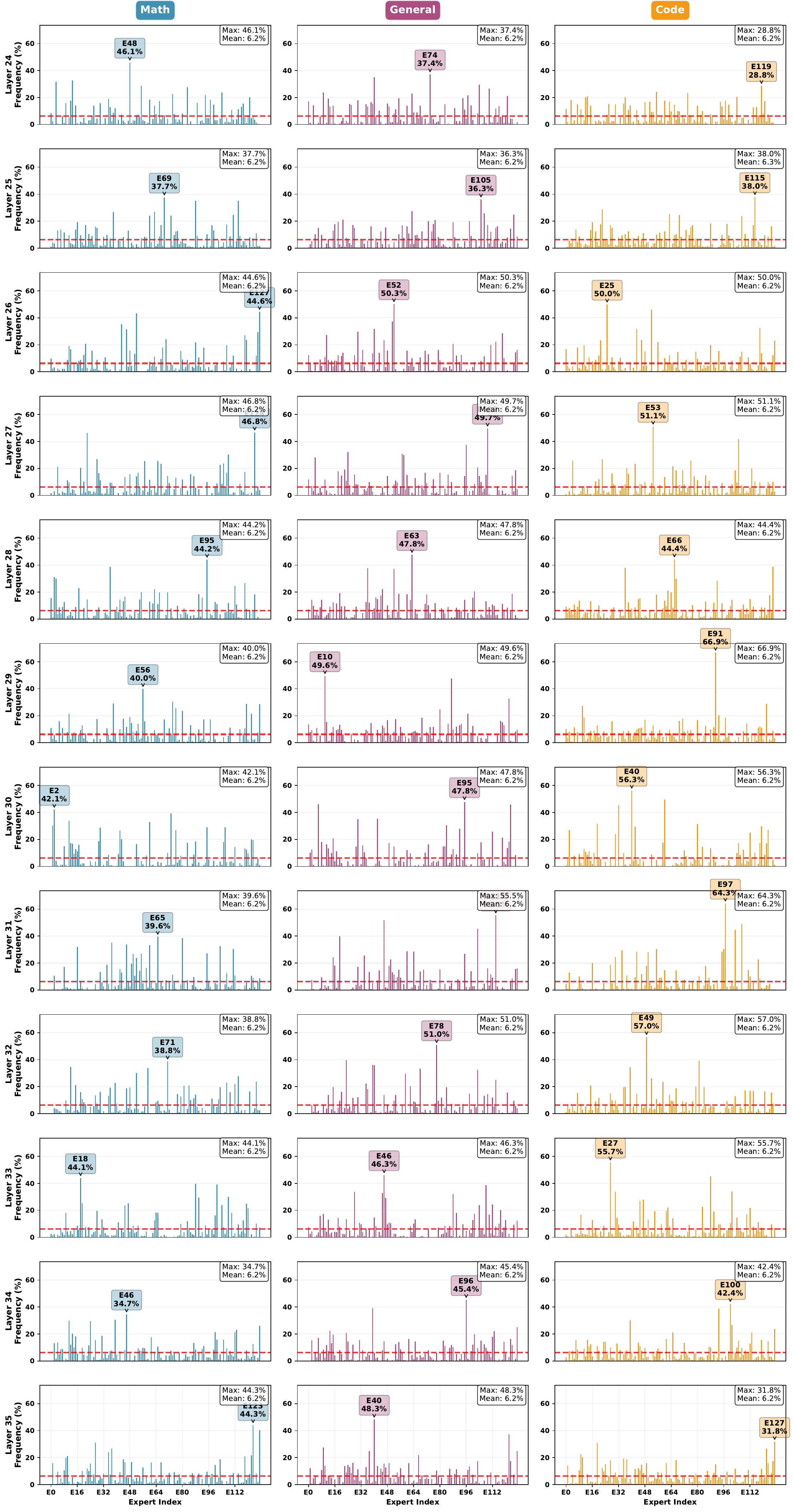}
    \caption{Expert usage frequency for layers 24--35 across math, code, and general tasks.}
    \label{fig:freq-24-35}
\end{figure*}

\begin{figure*}[t]
    \centering
    \includegraphics[height=0.95\textheight]{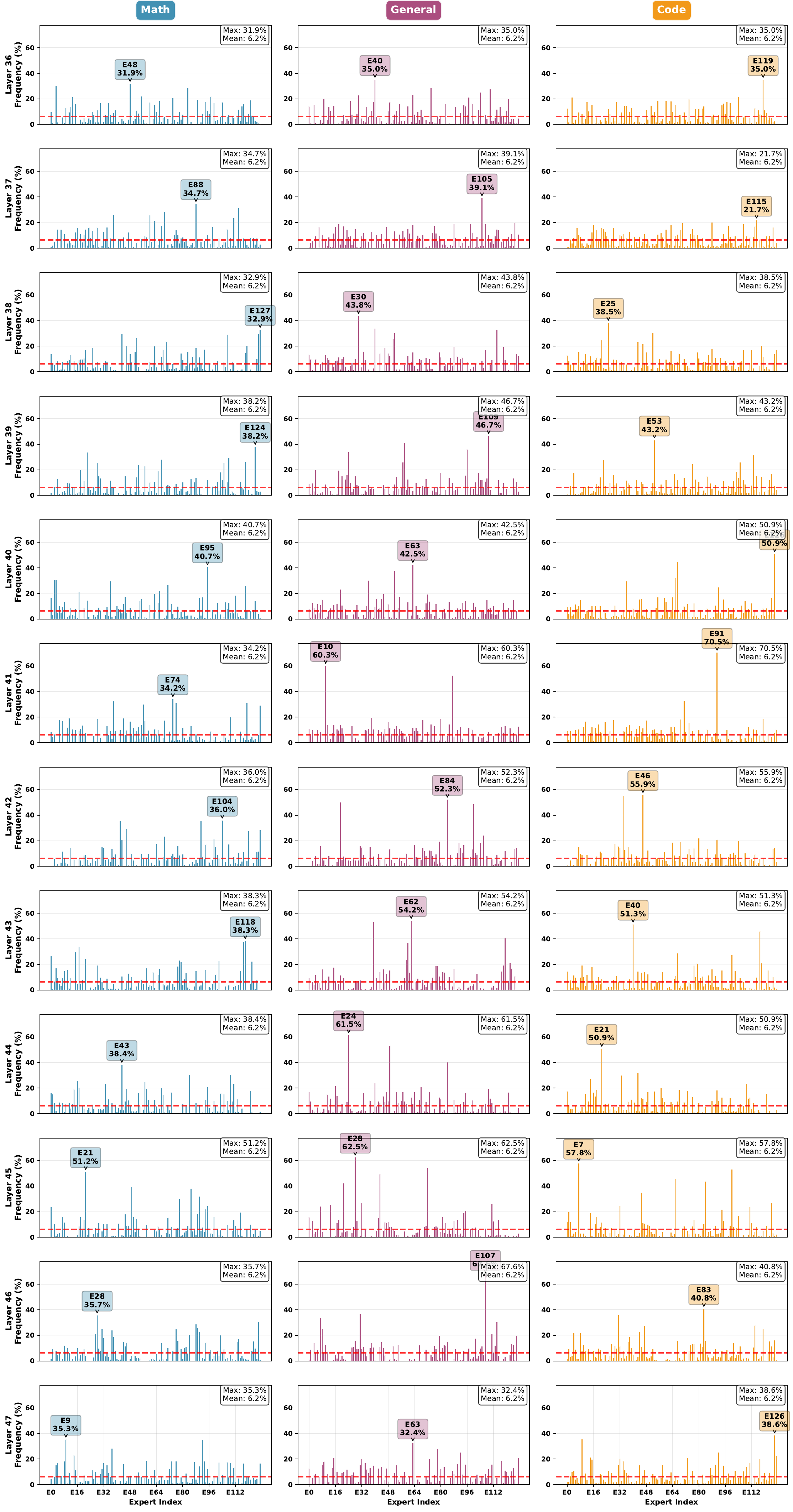}
    \caption{Expert usage frequency for layers 36--47 across math, code, and general tasks.}
    \label{fig:freq-36-47}
\end{figure*}

In Section~\ref{sec:expert-specialization}, we analyzed the specialization of experts in fine-grained MoE models using Qwen3-30B-A3B as a case study, and presented representative figures for a subset of layers. Here, we provide the complete visualization of expert usage frequencies across all layers of the model. Results are shown for three representative task types---math, code, and general. For each layer, the bar charts display the relative usage frequency of each expert, aggregated over all tokens in both the prefill and decode stages. Figures are grouped every 12 layers for readability. These results (Figures~\ref{fig:freq-0-11}--\ref{fig:freq-36-47}) confirm that different task types consistently activate distinct subsets of experts, highlighting clear patterns of domain specialization.

\section{Hyperparameter Choices in Dynamic Pruning}
\label{app:pruning_param}
As introduced in Section~\ref{sec:dynamic-pruning}, our dynamic pruning strategy involves two quantities: $K_{\min}$ and $\lambda$. 
The value of $K_{\min}$ is determined empirically to serve as a safeguard to prevent selecting too few experts. 
In contrast, $\lambda$ acts as the single knob controlling pruning aggressiveness and presents a trade-off between efficiency and accuracy. 
We discuss their effects through detailed experiments below, demonstrating that both hyperparameters are robust and easy to set in practice.

\paragraph{Setting of $K_{\min}$.}  
In the main text, we noted that reducing the number of selected experts below a certain threshold causes the model to collapse entirely. Therefore, $K_{\min}$ must guarantee that even when both layer- and token-level sensitivities are low, the number of selected experts never falls below this safe lower bound.  

To study this effect, we take Qwen3-30B-A3B as an example and directly enforce the number of experts per token to be fixed between 2 and 7 (with $K_{\text{base}}=8$ as the default). Table~\ref{tab:kmin} reports the performance on five benchmarks. For efficiency, results on AIME2024 (except the baseline) are measured over a single run.  

\begin{table*}[tbhp]
\centering
\small
\caption{Performance of Qwen3-30B-A3B under different fixed numbers of selected experts.}
\label{tab:kmin}
\begin{tabular}{c|ccccc}
\toprule
TopK & AIME2024 & Math500 & GPQA & Livecodebench & HumanEval+ \\
\midrule
8 & 80.67 & 95.2 & 65.66 & 91.75 & 71.34 \\
7 & 76.67 & 95.0 & 65.66 & 91.25 & 71.95 \\
6 & 73.33 & 93.2 & 63.64 & 89.75 & 69.51 \\
5 & 70.00 & 88.2 & 59.60 & 87.00 & 66.46 \\
4 & 63.33 & 78.4 & 54.55 & 74.75 & 59.76 \\
3 & 40.00 & 55.8 & 44.95 & 57.50 & 43.29 \\
2 & 16.67 & 19.6 & 21.72 & 25.50 & 17.07 \\
\bottomrule
\end{tabular}
\end{table*}

We observe that removing just one expert (TopK=7) leads to almost no performance degradation, consistent with the expert pruning observations reported in Section~\ref{sec:motivation} in the main text. When TopK falls between 6 and 4, some performance degradation becomes apparent, but the model's core reasoning capabilities remain largely intact—far from a performance collapse. The critical boundary occurs at TopK=3: performance collapses dramatically when fewer than three experts are selected. Therefore, we set $K_{\min}=3$ to prevent the model from falling below this threshold. On the DeepSeek series, we observed a similar phenomenon, further reinforcing this choice. Notably, $K_{\min}$ is not intended as a finely tuned optimal value, but rather as a fixed safe lower bound specific to a given fine-grained MoE architecture. Once determined, it requires no further tuning across different tasks or datasets.

\paragraph{Choice of $\lambda$.}  
The hyperparameter $\lambda$ controls the trade-off between accuracy preservation and inference speedup: larger values of $\lambda$ bias the pruning strategy toward retaining more experts (thus preserving accuracy), while smaller values encourage more aggressive pruning (thus improving throughput). To study this trade-off, we take Qwen3-30B-A3B as an example and vary $\lambda$ from $0.5$ to $0.9$ in increments of $0.1$. For each setting, we evaluate accuracy on the same five benchmarks as in the main experiments and measure the inference speedup using \texttt{vLLM} as the serving backend. The speedup is calculated as the ratio between the total inference time across all datasets and the baseline time without pruning. All results are reported from a single run.  

\begin{table*}[tbhp]
\centering
\small
\caption{Accuracy and inference speedup of Qwen3-30B-A3B under different values of $\lambda$.}
\label{tab:lambda}
\begin{tabular}{|c|c|c|c|c|c|c|c|}
\hline
$\lambda$ & Avg TopK & AIME2024 & Math500 & GPQA & Livecodebench & HumanEval+ & Speedup \\
\hline
    & 8    & 80.67 & 95.2 & 65.66 & 91.75 & 71.34 & 1.00 \\
\hline
0.5  & 4.33 & 73.33 & 93.8 & 59.60 & 86.50 & 65.85 & 1.30 \\
0.6  & 4.61 & 76.67 & 94.0 & 63.64 & 89.25 & 68.90 & 1.26 \\
0.7  & 4.82 & 80.00 & 94.6 & 64.65 & 90.00 & 70.12 & 1.25 \\
0.8  & 5.09 & 80.00 & 94.0 & 64.65 & 90.75 & 70.12 & 1.22 \\
0.9  & 5.32 & 80.00 & 94.8 & 64.14 & 90.50 & 70.73 & 1.20 \\
\hline
\end{tabular}
\end{table*}

From Table~\ref{tab:lambda}, we observe that when $\lambda$ is set to an extremely small value (e.g., 0.5 or 0.6), pruning becomes overly aggressive—achieving higher speedups but at the cost of noticeable performance degradation. However, when $\lambda \geq 0.7$, the model achieves significant inference acceleration with minimal accuracy loss. Moreover, within this range, $\lambda$ exhibits robust behavior: varying it from 0.7 to 0.9 does not substantially affect either accuracy or speedup. This suggests that as long as $\lambda$ is not set to an extremely small value, it is a robust and easy-to-set hyperparameter. In practice, we recommend $\lambda = 0.7$ as a default choice that balances efficiency and accuracy well.

\section{More Results on Broader Task Categories}
\label{app:more_results}

To further verify the universality and real-world applicability of our proposed method, we conduct additional evaluations on tasks beyond the domains covered in the main experiments. Specifically, we apply \textbf{Pick} to Qwen3-30B-A3B with only the ``general'' domain key experts enhanced, and evaluate on two held-out benchmarks: \textbf{C-Eval}~\citep{huang2023ceval}, a comprehensive Chinese cross-disciplinary benchmark spanning 52 diverse categories ranging from humanities to STEM; and \textbf{IFEval}~\citep{zhou2023instructionfollowingevaluationlargelanguage}, an instruction-following benchmark that is critical to real-world user experience.

As shown in Table~\ref{tab:more_results}, Pick consistently improves accuracy on both benchmarks, achieving a gain of 1.74\% on C-Eval and 1.85\% on IFEval. These results demonstrate that Pick generalizes well to a broad range of real-world tasks beyond the originally tested domains, including multi-disciplinary knowledge reasoning and instruction-following scenarios. This further confirms the universality of leveraging key experts to enhance MoE model performance.

\begin{table}[h]
    \centering
    \caption{Results of \textbf{Pick} on broader task categories using Qwen3-30B-A3B with only the ``general'' domain key experts enhanced.}
    \label{tab:more_results}
    \begin{tabular}{|c|c|c|}
    \hline
    Benchmark & Method & Accuracy (\%) \\ \hline
    \multirow{2}{*}{C-Eval}  & Baseline & 86.60 \\ \cline{2-3}
                             & + Pick & \textbf{88.34} \\ \hline
    \multirow{2}{*}{IFEval}  & Baseline & 85.95 \\ \cline{2-3}
                             & + Pick & \textbf{87.80} \\ \hline
    \end{tabular}
\end{table}

\section{On the Generality of the ``General Reasoning'' Domain}
\label{app:general_domain}

We further clarify that the ``general reasoning'' domain in our paper is in fact a broad task class rather than a narrow category. As illustrated by the word cloud in Figure~\ref{fig:expert_special}, general-domain key experts are associated with branch tokens useful for generic reasoning, which benefits a wide range of downstream tasks.

Consistent with this observation, Table~\ref{tab:interaction-results} shows that enhancing general-domain key experts alone not only improves GPQA-Diamond by 3.53\%, but also yields gains on AIME24 (+1.33\%, math) and HumanEval+ (+1.22\%, code), even though these latter benchmarks belong to specialized domains. This indicates general-domain key experts capture transferable reasoning capabilities that complement domain-specific experts.

Moreover, the additional results in Appendix~\ref{app:more_results} further reinforce this point: by enhancing only the general-domain key experts, Pick achieves consistent improvements on both C-Eval (a cross-disciplinary knowledge benchmark) and IFEval (an instruction-following benchmark), neither of which is tied to a single specialized domain. This demonstrates that the general-domain key experts indeed serve as a broad backbone for diverse reasoning scenarios.

In practice, this offers a convenient deployment strategy: if one does not target a specific domain such as math or code, simply enhancing general-domain key experts can provide broad improvements across heterogeneous tasks without requiring domain-specific tuning.

\section{Additional Empirical Evidence for Expert Utilization Inefficiencies}
\label{app:more_motivation}

To further support the empirical observations presented in Section~\ref{sec:motivation} (expert utilization inefficiencies and optimization potential), we provide additional results on a broader set of benchmarks. During early exploration, we conducted both expert amplification and pruning experiments on multiple datasets, including ARC-Challenge~\citep{allenai:arc}, HellaSwag~\citep{zellers2019hellaswag}, and OpenbookQA~\citep{OpenBookQA2018}, and observed consistent conclusions. These early-stage findings motivated us to further investigate and ultimately validate our method on more complex reasoning tasks. Winogrande was selected as the illustrative example in the main text because it is widely used in evaluation and easy to reproduce.

Here, we report the supplementary results on ARC-Challenge, HellaSwag, and OpenbookQA using Deepseek-V2-Lite-Chat. We apply the same set of amplified key experts (L1E43, L10E33, L25E11) identified on Winogrande, as these benchmarks share a similar ``general knowledge'' nature, and likewise reduce expert selection from 6 to 5 in the pruning experiment.

\begin{table}[h]
    \centering
    \caption{Supplementary results to support the observations in Section~\ref{sec:motivation}. ``+ Key Expert'' denotes jointly enhancing the three identified key experts, while ``Pruned (K=5)'' reduces the number of selected experts per token from 6 to 5.}
    \label{tab:more_motivation}
    \resizebox{\columnwidth}{!}{
    \begin{tabular}{|c|c|c|c|}
    \hline
    Dataset & Baseline & + Key Expert & Pruned (K=5) \\ \hline
    ARC-Challenge & 54.44 & \textbf{56.61} & 54.05 \\ \hline
    HellaSwag     & 62.44 & \textbf{64.47} & 62.25 \\ \hline
    OpenbookQA    & 45.60 & \textbf{47.40} & 45.80 \\ \hline
    \end{tabular}}
\end{table}

As shown in Table~\ref{tab:more_motivation}, the two key observations in Section~\ref{sec:motivation} consistently hold across all three additional benchmarks. First, jointly enhancing the identified key experts yields notable accuracy improvements (+2.17\% on ARC-Challenge, +2.03\% on HellaSwag, and +1.80\% on OpenbookQA), confirming that a small number of impactful experts are systematically underutilized by the original routing strategy. Second, reducing the number of activated experts from 6 to 5 incurs only marginal accuracy changes (within $\pm$0.4\%) while saving 16.7\% of expert computation, indicating that the redundancy in expert selection is a general phenomenon rather than a dataset-specific artifact. Together, these results reinforce that the expert utilization inefficiencies identified in the main paper are robust and broadly applicable, providing a solid empirical foundation for our proposed Ban\&Pick method.

\section{Mixture-of-Experts}
The Sparsely-Gated MoE \citep{6797059, 6796382} first introduced the idea of using a trainable gating network to activate only a small subset of experts, typically through a Top‑K routing scheme. Subsequent studies \citep{deepseekai2024deepseekv2strongeconomicalefficient, yang2025qwen3technicalreport} built upon this framework, scaling MoE models to much larger sizes and demonstrating that sparse expert activation can deliver strong efficiency and performance gains. As these models expanded, the routing mechanism became a critical factor in determining overall effectiveness.

The Sparsely-Gated MoE are based on a transformer architecture, but the FeedForward Network (FFN) sub-layers of traditional dense models are replaced with MoE layers, each containing $N$ experts. For each input token $\bm{x}$, the router computes routing logits \( \bm{r} = \{r_0, \cdots, r_{N-1}\} \) and expert selection scores \( \bm{s} = \text{Softmax}(\bm{r}) \). The top-$K$ experts are selected based on \( \bm{s} \), and their outputs \( E_{e_j}(\bm{x}) \) are combined as a weighted sum, with normalized weights:
\begin{equation}
\label{eq:moe_output}
\bm{z} = \sum_{j=0}^{K-1} \frac{\bm{s}_{e_j}}{\sum_{i=0}^{K-1} \bm{s}_{e_i}} \cdot E_{e_j}(\bm{x}).
\end{equation}
Here, \( E_{e_j}(\bm{x}) \) represents the output of the \( j \)-th selected expert for the input token \( \bm{x} \). Notably, different MoE architectures handle expert weight combination differently. Qwen3 employs the weight renormalization mechanism shown in Equation~\ref{eq:moe_output}, where the selected expert weights are normalized to sum to 1. In contrast, DeepSeek-v2 uses direct weight scaling without renormalization. Moreover, DeepSeek‑v2 incorporates grouped expert selection, where experts are partitioned into groups to improve specialization. Detailed settings and implementation can be checked in their papers \citep{deepseekai2024deepseekv2strongeconomicalefficient}.

Despite such architectural differences, both lines of work--and many other recent systems--adopt a fine‑grained MoE design: each layer hosts a large number of small, independently initialized experts. This design enhances expert specialization but simultaneously increases the difficulty of ensuring efficient routing. In our experimental evaluation, we examine four representative fine-grained MoE models, whose detailed architectural configurations and parameter settings are summarized in Table~\ref{tab:moe_specs}. Note that in the DeepSeek-v2 architecture design (including Deepseek-V2-Lite-Chat and Deepseek-v2.5-1210), the first layer is a dense layer for training stability, while the remaining layers are MoE layers. In contrast, the Qwen3 series uses MoE layers throughout all layers.

\begin{table*}[t]
\centering
\caption{Specifications of fine-grained MoE models used in our experiments.}
\label{tab:moe_specs}
\begin{tabular}{lcccccr}
\toprule
\textbf{Model} & \textbf{Experts} & \textbf{Top-K} & \textbf{Shared-Experts} & \textbf{Layers} & \textbf{Parameters} \\
\midrule
Deepseek-V2-Lite-Chat & 64 & 6 & 2 & 27 & 16B \\
Deepseek-v2.5-1210 & 160 & 6 & 2 & 60 & 236B \\
Qwen3-30B-A3B & 128 & 8 & 0 & 48 & 30B \\
Qwen3-235B-A22B & 128 & 8 & 0 & 94 & 235B \\
\bottomrule
\end{tabular}
\end{table*}

\section{Evaluation Details}
\label{app:evaluation}
\subsection{Dataset Evaluation.}
For \textbf{AIME2024}, due to the small sample size, we follow the common practice in technical reports by evaluating each method five times and reporting the average. For other datasets, constrained by time, we conduct only a single evaluation. On the \textbf{GPQA-Diamond} dataset, to further reduce errors caused by misaligned answer formats, we adopt an LLM-judge in addition to the base evaluator: specifically, we deploy a \texttt{Qwen2.5-72B-Instruct} model \citep{qwen2025qwen25technicalreport} with \texttt{vLLM} \citep{kwon2023efficientmemorymanagementlarge} as the judge model. For all other datasets, we rely on the default base evaluators. 

\subsection{Generation Hyperparameters.} 
During generation, for all four models used in our experiments, we adopt the recommended hyperparameters provided in their papers or Model Cards. The configuration is summarized in Table~\ref{tab:gen-hparams}.

\begin{table*}[h]
    \centering
    \caption{Generation hyperparameters for the evaluated models.}
    \label{tab:gen-hparams}
    \vskip 0.1in
    \begin{tabular}{lcccc}
        \toprule
        \textbf{Model} & \textbf{Top-k} & \textbf{Top-p} & \textbf{Temperature} & \textbf{Max Output Length} \\
        \midrule
        DeepSeek-V2-Lite-Chat & None & 0.95 & 0.3 & 8192 \\
        DeepSeek-V2.5-1210    & 1 & 0.90 & 0.3 & 8192 \\
        Qwen3-30B-A3B         & 20 & 0.95 & 0.6 & 32768 \\
        Qwen3-235B-A22B       & 20 & 0.95 & 0.6 & 32768 \\
        \bottomrule
    \end{tabular}
\end{table*}











\section{Details of Compared Methods and Reproduction Settings}
\label{app:reproduce}
In the main experiments, we compared our proposed \textsc{Ban} and \textsc{Pick} strategies against several existing baselines. For brevity, the main text only provided a concise overview of these methods. In this appendix, we present a more detailed description of each compared method, together with the specific reproduction settings used in our experiments. We organize the discussion into two parts: methods compared with \textsc{Pick}, and methods compared with \textsc{Ban}.  

\subsection{Compared Methods for \textsc{Pick}}
\paragraph{Dynamic Routing.}  
\citep{huang-etal-2024-harder} first proposed the idea of dynamically allocating experts, arguing that assigning a fixed number of experts to each token is overly rigid. Instead, the number of experts should be adapted according to the difficulty of the underlying task: tokens from challenging tasks should be allocated more experts, while tokens from easier tasks can be served with fewer. Concretely, experts are sequentially selected in descending order of routing weight until the cumulative weight exceeds a predefined threshold $\tau$. Formally, let $r_{(1)}, r_{(2)}, \dots, r_{(K)}$ denote the routing weights of experts sorted in descending order. The number of selected experts $k$ is determined as
\[
k = \min \Bigl\{ m \,\big|\, \sum_{j=1}^{m} r_{(j)} \geq \tau \Bigr\}.
\]
In our experiments, since the optimal threshold $\tau$ cannot be known in advance, we evaluate three candidate values $\tau \in \{0.7, 0.8, 0.9\}$ and report in the main text the result corresponding to the setting that achieves the best average accuracy across benchmarks.

\paragraph{TiP.}  
The core idea of TiP is motivated by the observation that, during chain-of-thought (CoT) reasoning, models often switch reasoning paths too frequently in the early exploration stage. Such premature abandonment of a correct path is referred to as \emph{underthinking}. To mitigate this issue, TiP introduces a penalty on tokens that typically indicate a switch of reasoning path (e.g., ``alternatively''). Concretely, after the softmax operation, the probability mass of these tokens is reduced by subtracting a fixed penalty value from their logits, thereby encouraging the model to continue deepening its current reasoning path rather than switching too early.  

This method involves two hyperparameters: $\alpha$ (penalty strength) and $\beta$ (penalty duration). The original paper did not provide an automatic tuning strategy but instead conducted a grid search to identify relatively effective values. Following their reported settings, we set $\alpha=3$ and $\beta=600$ in our reproduction.

\paragraph{Rice.}  
Rice builds on the observation that recent MoE-LLMs often employ an explicit \texttt{<think>} token to train chain-of-thought (CoT) reasoning ability. For example, the Qwen3 series can switch between deep-thinking and non-thinking modes by toggling the use of the \texttt{<think>} token. Rice leverages this property by comparing expert activation frequencies with and without the \texttt{<think>} token, thereby identifying cognitive experts that are closely associated with deep reasoning. Once identified, the routing weight of a cognitive expert is multiplied by a constant factor greater than $1$ whenever it is selected, amplifying its influence and enhancing the model’s reasoning ability.  

Since this method relies on explicit \texttt{<think>} tokens, it is only applicable to models such as the Qwen3 series, but not to other models used in our experiments. For reproduction, we followed the original paper’s settings: for Qwen3-235B-A22B, we directly used the cognitive experts reported therein; for Qwen3-30B-A3B, as the original paper did not list them, we applied the ICE (Identify Cognitive Experts) procedure proposed in the paper to locate the corresponding cognitive experts before applying the enhancement.  

\subsection{Compared Methods for \textsc{Ban}}
\paragraph{DES.}  
DES (Dynamic Expert Skip) was proposed to dynamically adjust the number of active experts by exploiting the observation that, for tokens that are relatively easy to process, the routing weights of the selected experts often differ significantly. In such cases, the contribution of the lower-weighted expert to the final output is negligible. Based on this, DES introduces a dynamic pruning strategy: whether to keep the lower-weighted expert is decided according to the ratio between expert weights. For example, in Mixtral-8$\times$7B \citep{jiang2024mixtral} where each token selects two experts, DES first computes the median ratio of top-1 to top-2 expert weights on a calibration set. During inference, if the actual ratio exceeds this median, the top-2 expert is skipped.  

In our reproduction, we aimed to align the pruning degree and acceleration ratio with our proposed Ban method, so that the comparison mainly reflects differences in accuracy preservation. For Qwen3 models, when $\lambda=0.7$, the average number of selected experts under Ban is slightly below 5. Therefore, we set the maximum number of experts to 6 and the minimum to 4. Specifically, we computed the medians of the ratios $\tfrac{r_{(4)}}{r_{(5)}}$ and $\tfrac{r_{(5)}}{r_{(6)}}$ on a calibration set, where $r_{(j)}$ denotes the $j$-th largest routing weight. At inference time, if the first ratio exceeds its median, only the top-4 experts are kept; otherwise, if the first ratio does not exceed its median but the second does, the top-5 experts are kept; otherwise, all top-6 experts are retained. For the DeepSeek models, we followed an analogous procedure to ensure comparable pruning degree and speedup.

\paragraph{ODP.}  
ODP (Online Dynamic Pruning) is an extension of DES that incorporates an additional key-token protection mechanism. While DES prunes experts dynamically based on weight ratios, ODP further observes that certain tokens carry disproportionately high importance in the attention map. If a token receives an attention score significantly larger than the others, pruning its experts may harm model performance. To address this, ODP protects such key tokens: even if their expert weights satisfy the pruning condition, they are exempted from pruning.  

In our reproduction, we followed exactly the same procedure as DES for the pruning part. On top of that, we adopted the evaluation metric for token importance proposed in the ODP paper to identify key tokens. Whenever a token is classified as key, we enforced the selection of all chosen experts for it, regardless of the pruning condition.

\section{Discussion on Strategy Selection for Pick Enhancement}

In the main text (Section~4.3), we showed through experiments on \textsc{MATH‑500}, \textsc{AIME2024}, and \textsc{AIME2025} that among the five proposed enhancement strategies for key experts, \textbf{Strategy D} achieves the most consistent accuracy improvements. Here, we provide a more detailed explanation of the motivations behind the design of these strategies and the reasoning that led us to ultimately adopt Strategy D.

\paragraph{Initial motivation: Strategies A and B.}
An initial intuition was straightforward—since key experts play a decisive role in task‑specific reasoning, one might expect model performance to improve if all tokens are explicitly routed through or enhanced by these key experts. Based on this idea, \textbf{Strategies A} and \textbf{B} were designed to directly enforce the participation of key experts for all tokens, aiming to strengthen their influence across the entire sequence regardless of token type or context. This approach represents the most direct attempt to leverage key experts’ strong impact on output generation to achieve better overall results.

\paragraph{Why Strategies A and B are not stable.}
However, subsequent analyses in Sections~4.1 and~4.2 of the main paper suggest that key experts are closely tied to task‑relevant tokens and can be regarded as a subset of domain‑specialized experts. Taking the \textit{math} domain as an example, these key experts are highly activated by digits, letters, and operators, but not by unrelated tokens such as pronouns or prepositions. Consequently, forcing key experts to be enhanced for every token disregards this natural specialization and leads to unstable or degraded performance across datasets. This behavior is \textbf{consistent with the experimental results in Section~4.3}, where Strategies A and B exhibit larger performance fluctuations and reduced accuracy compared with other strategies. These observations confirm that a more adaptive mechanism is needed to selectively enhance key experts where they are most beneficial, rather than applying uniform enforcement.

\begin{table*}[tbh]
\centering
\small
\caption{Routing frequency (\%) of the math key expert $(2,37)$ on the math calibration set. Tokens are taken from the \textit{math} and \textit{general} word clouds in Figure \ref{fig:expert_special}. Each column shows the proportion of tokens whose key expert appears in different routing weight ranges.}
\label{tab:keyexpert_top2k_distribution}
\begin{tabular}{lccc}
\toprule
\textbf{Token Type (20 tokens each)} & \textbf{Top-$k$ (\%)} & \textbf{Top-$2k$ but not Top-$k$ (\%)} & \textbf{Outside Top-$2k$ (\%)} \\
\midrule
Math tokens & 68.4 & 24.7 & 6.9 \\
General tokens & 2.5 & 0.4 & 97.1 \\
\bottomrule
\end{tabular}
\end{table*}

\paragraph{Adaptive Enhancement Yields Better Performance.}
To address the instability observed in Strategies A and B, we designed adaptive strategies that selectively enhance key experts based on token‑level routing signals rather than enforcing fixed participation. A natural idea might be to construct an explicit lookup table of task‑relevant tokens so that enhancement can be applied in a perfectly fine‑grained manner. However, such an approach would be overly cumbersome in implementation and inconsistent with our design objective of maintaining simplicity and efficiency. Instead, we empirically observed that for tokens within a key expert’s corresponding domain—for example, math‑related tokens—the key experts often appear within the top‑$2k$ routing weight range even when they are not selected in the top‑$k$ ($k$ is the default number of experts per token), whereas for unrelated tokens they seldom appear within the top‑$2k$. This observation provides a lightweight and effective criterion for identifying which tokens should trigger the enhancement of their domain‑specific experts.

We conducted this verification on \textsc{Qwen3‑30B‑A3B} using the math key expert $(2,37)$. All tests were performed on the \textit{math calibration subset}. To analyze cross‑domain behavior, we selected the top‑20 most frequent tokens from both the \textit{math} and \textit{general} word clouds shown in Figure \ref{fig:expert_special} of the main paper. For each token, we measured the frequency with which the math key expert appears in the top‑$k$, in the top‑$2k$ but not in the top‑$k$, and outside the top‑$2k$ routing weight range. The results are presented in Table \ref{tab:keyexpert_top2k_distribution}.

The results show that, even when evaluated purely on the math calibration set, the math key expert $(2,37)$ is highly activated for math‑related tokens—appearing in the top‑$2k$ routing range for more than 90\% of them—while rarely activated for general tokens. This indicates that the routing mechanism naturally captures a strong correlation between math tokens and the math key expert, validating that the adaptive strategy can precisely align experts with domain‑relevant tokens. Consequently, \textbf{Strategies C} and \textbf{D} leverage this intrinsic property to enhance key experts selectively, improving performance on relevant tokens without blind strengthening across the vocabulary.

\paragraph{Choice between Strategy C and D.}
The primary difference between Strategies C and D lies in whether the minimally weighted selected expert is replaced. Our observations indicate that the least‑weighted expert contributes negligibly to model performance, consistent with the motivation of the \textbf{Ban} module: pruning or replacing experts of minimal contribution improves efficiency with minimal accuracy loss. Thus, replacing or retaining this expert makes little difference; we therefore adopted \textbf{Strategy D} for its simplicity and implementation efficiency.

\paragraph{Future directions.}
We believe this analysis offers deeper insight into the design rationale of Pick. Future extensions may incorporate token‑adaptive or data‑driven policies to automatically determine enhancement conditions, potentially providing finer control and better synergy with the underlying Ban mechanism while maintaining the plug‑and‑play nature of the framework.

\begin{table}[tbh]
\centering
\small
\caption{System-level evaluation of Ban\&Pick on Qwen3-30B-A3B.}
\label{tab:throughput_memory}
\vspace{3pt}
\setlength{\tabcolsep}{4pt}
\begin{tabular}{lccc}
\toprule
\textbf{Metrics} & \textbf{Base} & \textbf{+Ban\&Pick} & \textbf{Change} \\
\midrule
Avg Top-$k$            & 8.00  & 4.82  & $-$39.75\% \\
Throughput (tokens/s)  & 10.01 & 12.85 & $+$28.37\% \\
Peak Memory (GB)       & 5.84  & 5.32  & $-$8.90\%  \\
\bottomrule
\end{tabular}
\end{table}

\section{System-Level Performance Evaluation}
\label{app:system_evaluation}

To conduct a more detailed system-level performance evaluation of our proposed \textbf{Ban\&Pick} method, we performed additional experiments on the \textbf{Qwen3‑30B‑A3B} model. The evaluation focuses on \textbf{throughput} and \textbf{memory usage}, which are important metrics for assessing the efficiency of large-scale models in real deployment scenarios.

All tests were executed on a single A800‑80G GPU. We used identical prompts and identical inference configurations for both the base model and the Ban\&Pick‑enhanced model (\textit{maximum output length} = 1k, two warm‑up runs, and five averaged runs). The average throughput and peak memory usage are summarized in Table~\ref{tab:throughput_memory}.

The results show that, due to expert pruning introduced by the Ban module, \textbf{throughput increases notably by +28.37\%}, while \textbf{peak memory usage decreases slightly}. These results demonstrate that Ban\&Pick not only improves model accuracy but also enhances \textbf{efficiency and deployability} in practical inference scenarios.

\section{LLM Usage}
In the preparation of this manuscript, large language models were used to polish the writing, including grammar refinement, phrasing improvement, and minor stylistic edits.

\end{document}